%% file: paper.tex
\documentclass[11pt,a4paper]{article}
\pdfoutput=1

\input{incl_settings.tex}
\input{incl_shortcuts.tex}

\input{tikz_setup}

\definecolor{titlecolor}{named}{black}
\usepackage[labelfont=bf,width=0.90\textwidth]{caption}

\begin{document}

\vspace*{-1.5cm}
\begin{flushright}
TIF-UNIMI-2026-11
\end{flushright}
\vspace{0.5cm}

\begin{center}
{\Large \textbf{\color{titlecolor}{
Unifying Generative Models with Path Integrals
}}}
\end{center}
\vspace{0.5cm}

\begin{center}
Ramon Winterhalder
\end{center}

\begin{center}
{}TIFLab, Dipartimento di Fisica, Universit\`a degli Studi di Milano,\\
and INFN, Sezione di Milano, Via Celoria 16, I-20133 Milano, Italy
\end{center}

\vspace{0.5cm}

\begin{center}
{\bf \large Abstract}
\end{center}
We formulate generative modeling as a path integral in which flow-based, diffusion-based, variational, and adversarial models arise as different evaluation principles for a single master action. Its Martin-Siggia-Rose-Janssen-de~Dominicis (MSRJD) form separates free from interacting probability flows and opens them to diagrammatic perturbation theory.
The expansion yields a one-loop correction to deterministic samplers at no stochastic-sampling cost, which we validate on solvable and nonlinear drifts, where it reduces a $53\,\%$ tree-level error to $1.6\,\%$.
Imperfect learned scores enter as insertions and yield a response-weighted score-matching objective, and symmetry-equivariant drift design becomes an operator expansion with EFT power counting.

\vspace{6pt}
\noindent\rule{\textwidth}{1pt}
{\setlength{\parskip}{0pt}\tableofcontents}
\noindent\rule{\textwidth}{1pt}

\clearpage
\section{Introduction}
\label{sec:intro}

Generative models sample complex, high-dimensional probability distributions fast and accurate enough to be relevant for physics, where probabilistic modeling carries simulation, inference, and uncertainty quantification.
In high-energy physics (HEP) they are being developed as accelerators for Monte Carlo event generation, detector simulation, and unfolding, while maintaining physical fidelity~\cite{Butter:2022rso, Plehn:2022ftl, Krause:2024avx}.
See the HEP-ML Living Review~\cite{Feickert:2021ajf,hepmllivingreview,krause_2026_21626667} and HEP-ML Living Guide~\cite{Krause:2026ayh} for a broader overview.

Generative models are typically introduced within their own mathematical language and training paradigm.
For instance, normalizing flows construct explicit invertible maps with tractable Jacobians, and diffusion models define generation as a time-reversed stochastic process.
From a physics perspective, this fragmentation is surprising.
Physics describes probabilistic dynamics through stochastic processes and path integrals, which provide a unified language for dynamics, fluctuations, and correlations and are routine tools in statistical mechanics and quantum field theory.
In particular, the Onsager--Machlup formulation~\cite{Onsager1953, Machlup1953} and the Martin-Siggia-Rose-Janssen-de~Dominicis (MSRJD) functional integral~\cite{Martin:1973zz, Janssen:1976qag, dominicis1976techniques} provide path-integral descriptions of stochastic dynamics that are tailor-made for the kind of probability transport generative models perform.

Prior work has already explored connections between subsets of these frameworks.
The score-based SDE framework unifies score matching and diffusion through stochastic differential equations running forward and backward in time~\cite{Song2021}, flow matching and stochastic-interpolant constructions provide deterministic training objectives for continuous-time transport~\cite{Lipman2023,Albergo2023}, and Refs.~\cite{DeBortoli2023, Berner2024} have developed the connections between diffusion models, optimal control, and Schrödinger bridges.
Path-integral and action-based viewpoints have appeared in Refs.~\cite{Nielsen2020, Verheyen:2022tov, huang2021, premkumar2023, hirono2024}.
Closest to our work, Ref.~\cite{hirono2024} formulates score-based diffusion models as path integrals and evaluates likelihood corrections in a semiclassical Wentzel--Kramers--Brillouin (WKB) expansion of an interpolating parameter connecting the probability-flow ODE to the reverse SDE.
No existing formalism does both. The works above unify subsets of these models, and Ref.~\cite{hirono2024} extracts corrections from a path integral, but not within a structure that also classifies the models it corrects.
Our construction differs in both scope and output. It embeds the full family of models above in one master action, its loop expansion corrects sample observables perturbatively without direct stochastic sampling, and it develops the operator side of the EFT analogy.

Starting from latent-variable models and their extension to chains of latent variables, we show that generative modeling can be formulated as a path integral over latent trajectories governed by an Onsager--Machlup action. 
Within this framework, normalizing flows, diffusion models, variational autoencoders, and generative adversarial networks emerge as distinct limits or approximations of a single master action, providing a classification scheme.
The classification is the starting point, but the reformulation also produces explicit results.
Once generative transport is a field theory, the error of a deterministic sampler, the effect of an imperfect score, and the design of a symmetry-respecting architecture become perturbative computations, with the same diagrammatic bookkeeping as corrections to a scattering process in quantum field theory.
The contributions of this work are:
\begin{enumerate}[label=(\roman*)]
\item We give a self-contained derivation that recovers normalizing flows, continuous normalizing flows, diffusion models, conditional flow matching, Schrödinger bridges, variational autoencoders, and generative adversarial networks as limits, special cases, or evaluation principles of a single Onsager--Machlup master action;

\item We recast this action in MSRJD form and identify the free/interacting split, its propagators, and a diagrammatic expansion of the transition kernel;

\item We use this structure to organize the discrepancy between a deterministic drift-based sampler and the full stochastic reverse process as a loop expansion in the diffusion strength. At one-loop order the expansion yields a covariance correction and a tadpole mean shift, obtained by integrating auxiliary equations alongside the deterministic trajectory. We compute these terms in closed form and validate them on an exactly solvable model and on nonlinear drifts;

\item We treat an imperfect learned score as a diagrammatic insertion and derive the corresponding response-weighted score-matching objective;

\item We show that EFT power counting, combined with a symmetry requirement, organizes the construction of equivariant drift architectures into a finite operator enumeration with a predicted coupling hierarchy.
    
\end{enumerate}
Throughout, the emphasis is theoretical, as we train no generative network. The numerical studies validate the loop formulas on synthetic drifts, where the exact answer is available.

The paper is organized as follows.
In \cref{sec:path_integral}, we formulate generative models as continuous probabilistic paths and derive the corresponding path-integral representation.
\Cref{sec:known_models} demonstrates how standard classes of generative models are recovered as limits of the master action.
In \cref{sec:qft}, we develop an analogy with statistical field theory, introducing free and interacting probability flows, a scattering interpretation, and the relation to the MSRJD formalism.
In \cref{sec:loop_corrections}, we apply this structure to derive and validate a loop expansion for the error of deterministic samplers.
In \cref{sec:eft_power_counting}, we use effective-field-theory power counting to construct symmetry-constrained probability flows.
We conclude with an outlook in \cref{sec:outlook}.

\clearpage
\section{Generative models as continuous probabilistic paths}
\label{sec:path_integral}

Latent-variable models and their extension to chains of latent variables lead to a path-integral representation.
Taking the continuum limit yields stochastic or deterministic dynamics in latent space, whose probability flow can be described by an effective action. This construction provides the common foundation on which different classes of generative models are built.

\subsection{From latent-variable models to path integrals}
\label{sec:from_latent_to_path}

Latent-variable models are among the most widely used probabilistic formulations of generative modeling~\cite{Bishop2006, Blei2017}.
In their simplest form, they express the data distribution $p_{\rm data}(x)$ as a marginal over latent degrees of freedom $z\in\mathbb{R}^d$,
\begin{align}
    p_{\rm data}(x) = \int \d z \; p(x|z)\,p_{\rm prior}(z)\eqcomma
    \label{eq:lvm}
\end{align}
where $p_{\rm prior}(z)$ denotes a prior distribution and $p(x|z)$ a conditional likelihood.
The central challenge is that the marginal $p_{\rm data}(x)$ is generally intractable for expressive choices of $p(x|z)$, since the integral over the latent space cannot be evaluated in closed form.
This also renders the true posterior,
\begin{align}
    p(z|x) = \frac{p(x|z)\,p_{\rm prior}(z)}{p_{\rm data}(x)}\eqcomma
    \label{eq:posterior}
\end{align}
intractable, as its normalization requires the very quantity $p_{\rm data}(x)$ one seeks to compute.
Variational autoencoders (VAEs)~\cite{Kingma2014,Kingma2019} address this by approximating the true posterior with a parametric inference network $q_\phi(z|x)$, trained to be as close as possible to $p(z|x)$.
Normalizing flows avoid this difficulty entirely by restricting to the deterministic special case $p(x|z)=\delta(x-\Phi(z))$ with an invertible map $\Phi$, for which we can evaluate the marginal exactly. We return to this in \cref{sec:known_models}.

A natural extension of \cref{eq:lvm} is to introduce a sequence of latent variables $\{z_k\}_{k=0}^N$ interpolating between a simple prior $p_{\rm prior}(z_N)$ and the data distribution $p_{\rm data}(z_0)\equiv p(x)$,
\begin{align}
    p_{\rm data}(z_0) 
    = \int \left[\prod_{k=1}^{N} \d z_k\; p(z_{k-1}|z_k)\right] p_{\rm prior}(z_N)\eqcomma
    \label{eq:lvm_chain}
\end{align}
which can be interpreted as a discrete path integral over latent trajectories. 
Here and in the following we identify the data variable with the terminal element of the chain, $z_0\equiv x$, writing $z_0$ whenever its role as the endpoint of a latent trajectory matters and keeping $x$ when we emphasize contact with the data.
Each transition kernel $p(z_{k-1}|z_k)$ defines one step of a Markov chain in the generative direction, and marginalization over intermediate states amounts to summing over all paths connecting the prior to the observation. 
Introducing the discrete action
\begin{align}
    S_N[z_{0:N}] = -\sum_{k=1}^N \log p(z_{k-1}|z_k)\eqcomma
    \label{eq:discrete_action}
\end{align}
we can write \cref{eq:lvm_chain} in the exponential form
\begin{align}
    p_{\rm data}(z_0) 
    = \int \Biggl[\prod_{k=1}^N \d z_k\Biggr]\, p_{\rm prior}(z_N)\,
      \exp\!\big[-S_N[z_{0:N}]\big]\eqcomma
    \label{eq:discrete_path_integral}
\end{align}
which illustrates the analogy to discrete path integrals from statistical mechanics~\cite{Kleinert:2009}.
At this stage, the action $S_N$ fully specifies the generative model through its transition probabilities.

\subsection{Continuum limit and the Onsager--Machlup action}
\label{sec:om_action}

We now take the continuum limit of the discrete latent chain in \cref{eq:discrete_path_integral}. 
Recall that the kernel $p(z_{k-1}|z_k)$ in the discrete action \labelcref{eq:discrete_action} runs in the \emph{generative} direction, \ie from prior ($k=N$) to data ($k=0$).
So far this kernel is arbitrary, and \cref{eq:discrete_action} is a rewriting valid for any Markov chain.
A continuum limit exists only once the one-step kernel is specified and carries a single power of $\Delta t$ such that the sum turns into a Riemann sum.
Diffusion processes provide such a kernel, which is why we define a forward SDE, invoke its time reversal as a known result, and discretize that reverse process to obtain $p(z_{k-1}|z_k)$ in closed form.
The continuum limit of the action then follows.

We set $\Delta t = 1/N$ and place the chain on the uniform grid $t_k = k\,\Delta t$ for $k=0,1,\ldots,N$, so that $t_0=0$ corresponds to data and $t_N=1$ corresponds to the prior.
From here on we write a grid state as $z(t_k)$ and reserve a subscript on $z$ for a time argument, so that $z_0\equiv z(0)$ and $z_1\equiv z(1)$ denote the data and prior endpoints.
Hence, we adopt the following convention throughout:
\begin{itemize}
    \item the \emph{forward} process runs from $t=0$ (data) to $t=1$ (prior),
    \item and the \emph{reverse} (generative) process runs from $t=1$ (prior) back to $t=0$ (data).
\end{itemize}
We start by modeling the forward process as a diffusion. Over a short interval $\Delta t$, the state evolves from the earlier state $z(t_{k-1})$ to the later state $z(t_{k})$ as
\begin{align}
    z(t_{k})
    = z(t_{k-1}) + f(z(t_{k-1}),t_{k-1})\,\Delta t + g(t_{k-1})\,\sqrt{\Delta t}\,\xi_{k}\eqcomma
    \qquad \xi_{k} \sim \mathcal{N}(0,\mathbb{I})\eqcomma
    \label{eq:continuum_scaling}
\end{align}
where $f(z,t)$ is the forward drift and $g(t)\geq 0$ a scalar diffusion coefficient.
\Cref{eq:continuum_scaling} is the Euler--Maruyama step of a diffusion process where the state moves deterministically by $f\,\Delta t$ and receives a Gaussian kick of size $g\,\sqrt{\Delta t}$, the $\sqrt{\Delta t}$ scaling being characteristic of Brownian noise.
A common choice is the variance-preserving process $f(z,t)=-\tfrac12\beta(t)\,z$ with $g(t)=\sqrt{\beta(t)}$, which drives any data distribution toward a standard-normal prior, while the variance-exploding process takes $f=0$ with a growing diffusion $g(t)$~\cite{Song2021}.
Taking $\Delta t\to 0$ defines the forward stochastic differential equation (SDE)
\begin{align}
    \d z = f(z,t)\,\d t + g(t)\,\d W_t\eqcomma
    \label{eq:sde}
\end{align}
with $W_t$ a standard Wiener process, \ie the continuous-time limit of a random walk~\cite{Gardiner1985}.
The induced density $p(z,t)$, with boundary conditions
\begin{align}
 p(z,t) \to 
 \begin{cases}
  p(z,0)\equiv p_{\rm data}(z)\qquad & t \to 0 \\
  p(z,1)\equiv p_{\rm prior}(z) \qquad & t \to 1  \eqcomma
\end{cases}
\label{eq:fm_limits}
\end{align}
evolves according to the Fokker--Planck equation~\cite{Risken1989, Gardiner1985}
\begin{align}
    \partial_t p(z,t)
    = -\nabla_z\cdot\big(f(z,t)\,p(z,t)\big)
      + \tfrac{1}{2}g^2(t)\,\nabla_z^2 p(z,t)\eqperiod
    \label{eq:fpe}
\end{align}
The generative model runs this process backward. The time reversal of the diffusion process in \cref{eq:sde} is itself a diffusion, satisfying the SDE~\cite{ANDERSON1982313, Haussmann1986TIMERO}
\begin{align}
    \d z = f_{\rm rev}(z,t)\,\d t + g(t)\,\d \bar W_t\eqcomma
    \qquad
    f_{\rm rev}(z,t)
    \equiv f(z,t) - g^2(t)\,\nabla_z\log p(z,t)\eqcomma
    \label{eq:rev_sde}
\end{align}
where $\bar W_t$ is a Wiener process run in the reverse direction and the reverse drift $f_{\rm rev}$ differs from the forward drift by the score $\nabla_z\log p(z,t)$ of the density.
The score term corrects the forward drift for the reversal of time, pointing toward regions of higher probability mass and counteracting the spreading induced by diffusion.

To obtain a path-integral representation, we discretize the reverse process.
On the grid $t_k = k\,\Delta t$, the standard Euler--Maruyama discretization of the reverse SDE~\cite{Maruyama:1955,Kloeden:1992,Higham:2001} expresses the earlier state $z(t_{k-1})$ in terms of the later state $z(t_k)$ as
\begin{align}
    z(t_{k-1})
    = z(t_k) - f_{\rm rev}(z(t_k),t_k)\,\Delta t
      + g(t_k)\,\sqrt{\Delta t}\,\xi_k\eqcomma
    \qquad \xi_k \sim \mathcal{N}(0,\mathbb{I})\eqcomma
    \label{eq:em_reverse_step}
\end{align}
the reverse-time counterpart of \cref{eq:continuum_scaling}, with the drift contributing $-f_{\rm rev}\,\Delta t$ as the step moves backward in $t$.
Equivalently, since the Euler--Maruyama step is linear in the Gaussian noise $\xi_k$, the transition is Gaussian in terms of the forward increment $\Delta z_k \equiv z(t_k) - z(t_{k-1})$,
\begin{align}
    p(z(t_{k-1}),\,t_{k-1} \mid z(t_k),\,t_k)
    &= \mathcal N_k\,
       \exp\!\left[
         -\frac{\|\Delta z_k -
                f_{\rm rev}(z(t_k),t_k)\,\Delta t\|^2}
               {2g^2(t_k)\,\Delta t}
       \right]\eqcomma\notag\\
    \mwith \qquad 
    \mathcal N_k &=\big(2\pi g^2(t_k)\,\Delta t\big)^{-d/2}\eqperiod
    \label{eq:short_time_kernel_rev}
\end{align}
At fixed $z(t_k)$, the densities in the earlier state $z(t_{k-1})$ and in the increment $\Delta z_k$ differ only by a constant shift with unit Jacobian, so the Gaussian in the increment is the conditional density itself.
This is exactly the transition kernel $p(z_{k-1}|z_k)$ entering the discrete chain in \cref{eq:lvm_chain}, now given an explicit Gaussian form by the reverse SDE.
Taking the logarithm,
\begin{align}
    -\log p(z(t_{k-1}),\,t_{k-1} \mid z(t_k),\,t_k)
    = \frac{\|\Delta z_k - 
             f_{\rm rev}(z(t_k),t_k)\,\Delta t\|^2}
           {2g^2(t_k)\,\Delta t}
      - \log \mathcal N_k\eqperiod
\end{align}
The second term is $z$-independent but we track it explicitly, as it requires care in the continuum limit.
For the first term, we introduce the forward velocity at step $k$ as
\begin{align}
\dot z(t_k) \equiv \frac{\Delta z_k}{\Delta t} 
= \frac{z(t_k) - z(t_{k-1})}{\Delta t}\eqcomma
\label{eq:forward_velocity}
\end{align}
so that
\begin{align}
    \frac{\|\Delta z_k - f_{\rm rev}(z(t_k),t_k)\,\Delta t\|^2}
         {2g^2(t_k)\,\Delta t}
    = \frac{\Delta t}{2g^2(t_k)}
      \|\dot z(t_k) - f_{\rm rev}(z(t_k),t_k)\|^2\eqperiod
\end{align}
Inserting into the discrete action gives a $z$-dependent sum together with
the collected Gaussian normalizations,
\begin{align}
S_N[z_{0:N}]
&= \sum_{k=1}^N
    \frac{\Delta t}{2g^2(t_k)}\,
    \bigl\|\dot z(t_k) - f_{\rm rev}(z(t_k),t_k)\bigr\|^2
  + \mathcal{C}_N\eqcomma
\notag\\
\mwith \qquad \mathcal{C}_N
&=-\sum_{k=1}^N \log\mathcal N_k\eqperiod
\label{eq:discrete_action_explicit}
\end{align}
The $z$-dependent sum carries a single power of $\Delta t$ per term and is thus a Riemann sum. 
Sending $N\to\infty$, \ie $\Delta t\to0$, it converges to the Onsager--Machlup action,
\begin{align}
    S_\text{OM}[z]
    = \int_0^1 \d t\;
      \frac{\|\dot z(t) - f_{\rm rev}(z(t),t)\|^2}{2g^2(t)}
    \equiv \int_0^1 \d t\;\mathcal L_\text{OM}\big(z(t),\dot z(t)\big)\eqcomma
    \label{eq:om_action}
\end{align}
where the last line defines the corresponding Onsager--Machlup Lagrangian $\mathcal L_\text{OM}$.
The $z$-independent term $\mathcal{C}_N$, by contrast, does \emph{not} converge and it diverges logarithmically as $\Delta t\to0$. 
This is the familiar situation for path integrals where the action and the measure do not have separate continuum limits, only their combination does. 
We therefore define the continuum path measure $\mathcal{D}z$ to carry the divergent normalization, integrating over the interior of the time grid only,
\begin{align}
    \mathcal{D}z
    \equiv
    \lim_{N\to\infty}\;
    e^{-\mathcal{C}_N}
    \prod_{k=1}^{N-1}\d z(t_k)\eqcomma
    \label{eq:path_measure_def}
\end{align}
so that the product $\mathcal{D}z\,e^{-S_\text{OM}[z]}$ is finite even though neither factor is. With this definition the continuum limit is taken for the path integral as a whole, not for the action in isolation.
The assignment of the finite parts between action and measure is itself a convention rather than uniquely fixed, \ie only the combination $\mathcal{D}z\,e^{-S_\text{OM}[z]}$ carries meaning. The same applies to the action itself, which is fixed only up to the discretization convention, cf.\ the It\^o--Stratonovich discussion below.
The endpoints $z(t_0)$ and $z(t_N)$ are deliberately excluded from the measure, as their values are held fixed and not integrated over.
The data endpoint $z_0 = z(t_0)$ is identified with $x$ as in \cref{sec:from_latent_to_path}, and $z_1 = z(t_N)$ is the prior endpoint.
This defines the transition kernel $K$ of the generative process as the continuum limit of the composed chain,
\begin{align}
K(z_0,0\mid z_1,1)
\equiv
\lim_{N\to\infty}
\int\Biggl[\prod_{k=1}^{N-1}\d z(t_k)\Biggr]
\prod_{k=1}^{N} p\big(z(t_{k-1})\mid z(t_k)\big)=
\int
\mathcal{D}z\;
\exp\!\left[
- S_\text{OM}[z]
\right]\eqcomma
\label{eq:kernel_endpoint_def}
\end{align}
where $K(z_0,0\mid z_1,1)$ is
the conditional probability of arriving at the data point $z_0$ at $t=0$ given the prior point $z_1$ at $t=1$. This is a pinned path integral summing over all latent trajectories connecting the two endpoints, weighted by the Onsager--Machlup action.
The boundary values of the path integral are fixed by the arguments of the kernel, $z(0)=z_0$ and $z(1)=z_1$, and by \cref{eq:path_measure_def} they are not part of the measure. 
We therefore suppress boundary conditions on the integral sign throughout.
The same construction on any subinterval of the grid, with the path integral running over the interior of $[t_i,t_f]$ and the values $z(t_i)=z_i$ and $z(t_f)=z_f$ fixed, yields the two-time kernel
\begin{align}
K(z_i,t_i\mid z_f,t_f)
=
\int
\mathcal D z\;
\exp\!\left[
-\int_{t_i}^{t_f} \d t\;\mathcal{L}_\text{OM}\big(z(t),\dot z(t)\big)
\right]
\qquad \mwith \quad t_i<t_f\eqcomma
\label{eq:scattering_kernel_def}
\end{align}
the transition density of the reverse process from the state $z_f$ at time $t_f$ to the state $z_i$ at time $t_i$, with \cref{eq:kernel_endpoint_def} the special case connecting prior and data.
Discretizing $[t_i,t_f]$ on a grid of $N$ steps with an interior seam at index $s$, the $N$ per-step normalizations combine into
$e^{-\mathcal C_N}$ while the $N-1$ integrations run over the interior slices only. 
Writing $\Delta t\,\mathcal L^{(k)}$ for the $k$-th summand of the discrete action in \cref{eq:discrete_action_explicit}, the discrete counterpart of $\mathcal L_\text{OM}$, the product factorizes at the seam.
Splitting the interior integrals there and taking the continuum limit in the two factors separately yields the exact composition law
\begin{align}
K(z_i,t_i\mid z_f,t_f)
&\equiv
\int
\mathcal D z\;
\exp\!\left[
-\int_{t_i}^{t_f} \d t\;\mathcal{L}_\text{OM}\big(z(t),\dot z(t)\big)
\right]\notag\\
&=
\lim_{N\to\infty}
\int \prod_{k=1}^{N-1}\d z(t_k)\;
\prod_{k=1}^{N}\mathcal N_k\,
e^{-\Delta t\,\mathcal L^{(k)}}\notag\\
&=
\lim_{N\to\infty}
\int \prod_{k=1}^{N-1}\d z(t_k)\;
\Bigg(\prod_{k=1}^{s}\mathcal N_k\,e^{-\Delta t\,\mathcal L^{(k)}}\Bigg)
\Bigg(\prod_{k=s+1}^{N}\mathcal N_k\,e^{-\Delta t\,\mathcal L^{(k)}}\Bigg)\notag\\
&=
\lim_{N\to\infty}
\int \d z_s\;
\underbrace{\int\!\prod_{k=1}^{s-1}\!\d z(t_k)\;\prod_{k=1}^{s}\mathcal N_k\,e^{-\Delta t\,\mathcal L^{(k)}}}
          _{\displaystyle \longrightarrow\; K(z_i,t_i\mid z_s,t_s)}\;
\underbrace{\int\!\prod_{k=s+1}^{N-1}\!\d z(t_k)\;\prod_{k=s+1}^{N}\mathcal N_k\,e^{-\Delta t\,\mathcal L^{(k)}}}
          _{\displaystyle \longrightarrow\; K(z_s,t_s\mid z_f,t_f)}\notag\\
&=
\int \d z_s\; K(z_i,t_i\mid z_s,t_s)\,K(z_s,t_s\mid z_f,t_f)
\qquad \mfor \qquad t_i<t_s<t_f\eqcomma
\label{eq:chapman_kolmogorov}
\end{align}
giving the Chapman--Kolmogorov equation~\cite{Gardiner2009,FeynmanHibbs1965}, the continuum remnant of the Markov property of the discrete chain.
As the transition density of the reverse diffusion, the kernel satisfies the associated Fokker--Planck equation in its first arguments~\cite{Risken1989,Gardiner1985},
\begin{align}
-\partial_t K(z,t\mid z',t')
&=
\nabla_z\cdot\Big(
f_{\mathrm{rev}}(z,t)\,
K(z,t\mid z',t')
\Big)
+
\tfrac12\,g^2(t)\,
\nabla_z^2 K(z,t\mid z',t')\eqcomma
\label{eq:kernel_fpe}
\\
K(z,t'\mid z',t')
&=
\delta(z-z')\eqperiod
\label{eq:kernel_terminal}
\end{align}
The first equation is written with respect to the sampling direction and is consistent with the forward equation~\labelcref{eq:fpe}. 
The delta-function boundary condition states that, before any evolution has taken place, a process initialized at $z'$ is localized at $z'$. 
It is also the vanishing-time limit of the one-step kernel in \cref{eq:short_time_kernel_rev}.
Integrating \cref{eq:kernel_fpe} over $z$, the right-hand side is a total divergence and vanishes for sufficiently rapid decay at infinity, such that
\begin{align}
\partial_t
\int \d z\;
K(z,t\mid z',t')=0\eqperiod
\end{align}
The normalization is therefore independent of $t$. The boundary condition in \cref{eq:kernel_terminal} gives
\begin{align}
\int \d z\;
K(z,t'\mid z',t')
=
\int \d z\;
\delta(z-z')=1\eqcomma
\end{align}
and consequently
\begin{align}
\int \d z\;
K(z,t\mid z',t')=1
\qquad
\mfor
\qquad
t\leq t'\eqperiod
\label{eq:kernel_normalization}
\end{align}
Thus, the reverse process initialized at $z'$ at time $t'$ reaches some state at the earlier time $t$ with unit probability. This is the continuum counterpart of the normalization of every Gaussian factor in \cref{eq:kernel_endpoint_def}.
Reinstating the prior weight and integrating over the prior endpoint gives the data density,
\begin{align}
p_{\rm data}(z_0)
=
\int \d z_1\; p_{\rm prior}(z_1)\,K(z_0,0\mid z_1,1)\eqcomma
\label{eq:scattering_density}
\end{align}
which is the latent-variable representation we started from in \cref{eq:lvm}, now with the conditional likelihood realized as a pinned path integral over the intervening trajectory, \ie
\begin{align}
K(z_0,0\mid z_1,1)\equiv p(z_0|z_1)\eqperiod 
\end{align}
Writing the kernel out, the data density takes the master path-integral form
\begin{align}
    p_{\rm data}(z_0)
    = \int \d z_1\; p_{\rm prior}(z_1)
    \int \mathcal{D}z\;
      \exp\!\left[
      -\int_0^1 \d t\;
      \frac{\|\dot z(t) - f_{\rm rev}(z(t),t)\|^2}{2g^2(t)}
      \right]\eqcomma
    \label{eq:path_integral_OM}
\end{align}
the prior-weighted sum over all latent trajectories ending at the data point, used repeatedly below.
Different choices of forward drift $f$ and diffusion coefficient $g(t)$ give rise to different $f_{\rm rev}$, and hence different generative models, as we show in \cref{sec:known_models}.

\subsubsection*{It\^o versus Stratonovich discretization}

We derived the action in \cref{eq:om_action} using the It\^o convention, which evaluates the drift at the later time 
$z(t_k)$. 
The Stratonovich convention instead uses the midpoint $\tfrac12[z(t_{k-1})+z(t_k)]$.
Expanding the midpoint drift about $z(t_k)$ and using $\langle\Delta z_k\rangle = f_{\mathrm{rev}}\Delta t$ generates an additional contribution of order $\mathcal{O}(\Delta t)$ proportional to the divergence of the drift, giving~\cite{Zinn-Justin:2002ecy, Aron:2010ac}
\begin{align}
    S_N^{\text{Strat}}[z]
    = S_N^{\text{It\^o}}[z]
      + \frac{1}{2}\int_0^1 \d t\;\nabla_z\cdot f_{\rm rev}(z,t)
      + \mathcal{O}(\Delta t^{1/2})\eqperiod
    \label{eq:ito_strat_difference}
\end{align}
The additional term is a local functional of $z$. It can shift the finite-noise stationarity equations, but it is of order $g^0$ against the Onsager--Machlup term of order $g^{-2}$ and therefore leaves the leading zero-noise trajectory unchanged.
In the It\^o discretization used throughout, the causal functional Jacobian is field independent, and we absorb it into the normalization of the path measure, while a Stratonovich or midpoint discretization would instead retain the divergence term as a local contribution to the action.

\subsection{Known generative models as limits of the master action}
\label{sec:known_models}

The path-integral formulation developed above provides a unified description of generative modeling as probability transport in latent space, governed by an effective action. 
Different generative architectures correspond to particular limits, parameterizations, or approximations of this master construction.
Normalizing flows, diffusion models, conditional flow matching, Schr\"odinger bridges, and variational autoencoders then follow as different ways of evaluating or approximating the same path integral, and adversarial models appear in the case where the saddle point admits no likelihood to evaluate.

\subsubsection{Normalizing flows as deterministic saddle points}
\label{sec:nf_saddle}

Normalizing flows arise as the deterministic saddle-point limit of the path-integral formulation, obtained by sending $g(t)\to 0$. In this limit the path integral is dominated by the single path on which the action is stationary, its saddle point, in the same way that classical mechanics emerges from the quantum path integral as $\hbar\to0$.
The normalizing flow density formula follows directly from the latent-variable representation
\begin{align}
    p_{\rm data}(x) = \int \d z\; p(x|z)\, p_{\rm prior}(z) = \int \d z\; \delta(x - \Phi(z))\, p_{\rm prior}(z)\eqcomma
\end{align}
where we have used that the generative kernel is deterministic.
For an invertible differentiable map $\Phi$ from latent to data space, $x = \Phi(z)$, the delta function
$\delta(x-\Phi(z))$ has a unique root at $z_\star=\Phi^{-1}(x)$. 
Applying the change-of-variables rule for delta functions,
\begin{align}
    \delta(x - \Phi(z)) = \frac{\delta(z - \Phi^{-1}(x))}{\left|\det \dfrac{\partial \Phi(z)}{\partial z}\right|_{z = \Phi^{-1}(x)}}\eqcomma
    \label{eq:change_of_variable_delta}
\end{align}
we obtain
\begin{align}
p_{\rm data}(x)
= \int \d z\; \delta(z - \Phi^{-1}(x))\frac{p_{\rm prior}(z)}{\left|\det \dfrac{\partial \Phi(z)}{\partial z}\right|_{z = \Phi^{-1}(x)}} = \frac{p_{\rm prior}\!\left(\Phi^{-1}(x)\right)}{\left|\det \dfrac{\partial \Phi(z)}{\partial z}\right|_{z = \Phi^{-1}(x)}}\eqperiod
\label{eq:nf_direct}
\end{align}
Using the inverse function theorem
\begin{align}
\dfrac{\partial \Phi^{-1}(x)}{\partial x} = \left.\left(\dfrac{\partial \Phi(z)}{\partial z}\right)^{-1}\right|_{z=\Phi^{-1}(x)}\eqcomma
\label{eq:ift}
\end{align}
the density takes the equivalent form
\begin{align}
    p_{\rm data}(x) = p_{\rm prior}\!\left(\Phi^{-1}(x)\right)\,\left|\det \frac{\partial \Phi^{-1}(x)}{\partial x}\right|\eqcomma
    \label{eq:nf_direct_inv}
\end{align}
which is exactly the standard change-of-variables formula for normalizing flows.

\subsubsection*{Functional delta distribution}

The path-integral derivation recovers this result as the $g \to 0$ saddle-point limit of the Onsager--Machlup action, and further generalizes it to the continuous-time setting of continuous normalizing flows (CNFs).
We introduce the equation-of-motion residual
\begin{align}
\mathcal{E}[z](t) \equiv \dot z(t)-f_{\rm rev}(z(t),t)\eqcomma
\label{eq:eom_residual}
\end{align}
so that the Onsager--Machlup weight of \cref{eq:om_action} reads
\begin{align}
e^{-S_\text{OM}[z]}
=
\exp\!\left[-\int_0^1\d t\;\frac{\|\mathcal{E}[z](t)\|^2}{2g^2(t)}\right]\eqperiod
\label{eq:om_weight}
\end{align}
As $g(t)\to0$, this weight is exponentially suppressed whenever $\mathcal{E}[z](t)\neq0$ for any $t$. For a fixed path $z$ with $\|\mathcal{E}[z]\|^2=\epsilon>0$, it vanishes as $\exp(-\epsilon/2g^2)\to0$.
The path integral therefore concentrates on the unique deterministic path satisfying the saddle condition
\begin{align}
\dot z_\star(t)
=
f_{\rm rev}(z_\star(t),t)
\qquad \mwith \qquad
z_\star(1)=z_1\eqcomma
\label{eq:nf_classical_path}
\end{align}
the generative trajectory launched from the prior sample $z_1$ at $t=1$ and arriving at the data point $z_\star(0)$ at $t=0$. 
As $g\to0$ the score contribution to the reverse drift vanishes, $f_{\rm rev}\to f$, so the saddle path is also governed by the ODE $\dot z_\star = f(z_\star,t)$ and depends deterministically on its endpoints.
Integrating it defines the generative flow map and its inverse,
\begin{align}
\Phi:\; z_1 \longmapsto z_\star(0)\quad\text{with}\quad z_\star(1)=z_1\eqcomma
\qquad
\Phi^{-1}:\; z_0 \longmapsto z_\star(1)\quad\text{with}\quad z_\star(0)=z_0\eqcomma
\label{eq:flow_map}
\end{align}
The notation is deliberate. 
Integrating the drift from the prior time to the data time realizes the invertible map $\Phi$ of \cref{eq:nf_direct}, so the discrete and continuous constructions share one symbol because they construct one object.
The required uniqueness and invertibility are guaranteed under the standard Picard--Lindel\"of assumptions, namely continuity in $t$ and local Lipschitz continuity in $z$, together with conditions ensuring existence on the full interval $t\in[0,1]$~\cite{CoddingtonLevinson1955,ArnoldODE,HaleODE}.

Since the normalization required for the Gaussian-to-delta limit is already carried by the continuum measure $\mathcal{D}z$, the deterministic limit turns the weighted measure into a functional delta, in the sense of distributions,
\begin{align}
\mathcal{D}z\;
\exp\!\left[-\int_0^1\d t\;\frac{\|\mathcal{E}[z](t)\|^2}{2g^2(t)}\right]
\;\xrightarrow{\;g\to0\;}\;
\mathcal{D}z\;\delta[\mathcal{E}[z]]\eqcomma
\label{eq:functional_delta_limit_nf}
\end{align}
enforcing the deterministic flow equation. 
To evaluate the pinned path integral we return to the discrete chain from \cref{eq:kernel_endpoint_def}, where the limit acts factor by factor.
Each Gaussian one-step kernel in \cref{eq:short_time_kernel_rev}
degenerates to a delta enforcing one Euler step of the flow,
\begin{align}
p\big(z(t_{k-1})\mid z(t_k)\big)
\;\xrightarrow{\;g\to0\;}\;
\delta\big(z(t_{k-1}) - z(t_k) + f(z(t_k),t_k)\,\Delta t\big)\eqperiod
\label{eq:nf_onestep_delta}
\end{align}
Each interior integration removes one delta with unit Jacobian, as the constraint is linear in the earlier state $z(t_{k-1})$, fixing the interior states successively along the classical trajectory from the prior end downward, $z(t_k)=z_\star(t_k)$.
After all interior integrations a single constraint remains, pinning the data endpoint to the arrival point of the flow,
\begin{align}
K(z_0,0\mid z_1,1)
\;\xrightarrow{\;g\to0\;}\;
\delta\big(z_0 - \Phi(z_1)\big)\eqperiod
\label{eq:nf_kernel_collapse}
\end{align}
In the deterministic limit the kernel degenerates to deterministic transport of the prior point along the flow. 
Inserting \cref{eq:nf_kernel_collapse} into \cref{eq:scattering_density} and resolving the ordinary delta with the same change-of-variables rule as in \cref{eq:nf_direct}, we obtain
\begin{align}
p_{\rm data}(z_0)
&=
\int \d z_1\; p_{\rm prior}(z_1)\,
\delta\big(z_0 - \Phi(z_1)\big)\notag\\
&\overset{\eqref{eq:change_of_variable_delta}}{=}
\int \d z_1\; p_{\rm prior}(z_1)\,
\frac{\delta\big(z_1 - \Phi^{-1}(z_0)\big)}
     {\left|\det\dfrac{\partial \Phi(z_1)}{\partial z_1}\right|_{z_1=\Phi^{-1}(z_0)}}\notag\\
&\overset{\eqref{eq:ift}}{=}
p_{\rm prior}\big(\Phi^{-1}(z_0)\big)\,
\left|\det\frac{\partial \Phi^{-1}(z_0)}{\partial z_0}\right|\eqperiod
\label{eq:nf_density_transform}
\end{align}
This is the normalizing-flow density formula, now with the continuous flow map $\Phi$ playing the role of the discrete invertible transform $f$.
Considering the trajectory as a function of its data endpoint $z_\star(0)=z_0$ and
differentiating the flow equation \labelcref{eq:nf_classical_path} with respect to $z_0$, we obtain the variational equation
\begin{align}
\frac{\d}{\d t}M(t)
=
\partial_z f(z_\star(t),t)\,M(t)
\qquad \mwith \qquad
M(t)
=
\frac{\partial z_\star(t)}{\partial z_0}
\qquad \mand \qquad
M(0)=\mathbb{I}\eqperiod
\label{eq:jacobian_ode}
\end{align}
The Jacobian entering \cref{eq:nf_density_transform} is
$M(1)=\partial z_\star(1)/\partial z_0=\partial\Phi^{-1}(z_0)/\partial z_0$.
Using Jacobi's formula,
\begin{align}
\frac{\d}{\d t}\log|\det M(t)|
=
\mathrm{tr}\left(
\frac{\d M(t)}{\d t}\,M(t)^{-1}
\right)\eqcomma
\end{align}
and inserting \cref{eq:jacobian_ode}, we thus obtain
\begin{align}
\frac{\d}{\d t}\log\bigl|\det M(t)\bigr|
=
\mathrm{tr}\left(
\partial_z f(z_\star(t),t)
\right)
=
\nabla_z\cdot f(z_\star(t),t)\eqperiod
\label{eq:liouville_jacobian}
\end{align}
Integrating from $t=0$ to $t=1$ with $\log|\det M(0)|=0$ gives
\begin{align}
\log\big|\det M(1)\big|
=
\int_0^1\d t\;
\nabla_z\cdot f(z_\star(t),t)\eqcomma
\end{align}
and substituting into \cref{eq:nf_density_transform} yields the continuous normalizing-flow log-likelihood,
\begin{align}
\log p_{\rm data}(z_0)
=
\log p_{\rm prior}\big(\Phi^{-1}(z_0)\big)
+
\int_0^1 \d t\;
\nabla_z \cdot f(z_\star(t),t)\eqcomma
\label{eq:nf_cnf_likelihood}
\end{align}
here derived directly from the saddle-point evaluation of the Onsager--Machlup path integral.

\subsubsection{Diffusion models as stochastic probability flows}
\label{sec:diffusion_as_score}

Diffusion-based generative models correspond to the regime in which stochasticity plays an essential role and the master path integral cannot be reduced to a single dominant trajectory.
Starting again from the master representation of the data density,
\begin{align}
p_{\rm data}(z_0)
=
\int \d z_1\; p_{\rm prior}(z_1)
\int \mathcal D z\;
\exp\!\left[
  -\int_0^1 \d t\;
  \frac{\|\dot z(t) - f_{\mathrm{rev}}(z(t),t)\|^2}{2g^2(t)}
\right]\eqcomma
\label{eq:diffusion_master}
\end{align}
we observe that for $g(t)\neq 0$ the functional weight does not collapse onto a single trajectory. Generative sampling now corresponds to drawing entire latent paths $z(t)$ from the path measure defined by the Onsager--Machlup action.
While \cref{eq:diffusion_master} provides a closed-form expression for $p_{\rm data}(z_0)$, it is intractable in practice. 
Evaluating it requires integrating over all trajectories \emph{and} knowing the reverse drift $f_{\mathrm{rev}}(z,t)$, which itself depends on the unknown score $\nabla_z\log p(z,t)$.
Diffusion models are therefore not trained on the data density directly. 
They rely on alternative evaluation principles that bypass the explicit computation of the path integral.

\subsubsection*{Score matching}

A common strategy is to specify a forward noising process with prescribed drift $f(z,t)$ and diffusion $g(t)$,
\begin{align}
    \d z = f(z,t)\,\d t + g(t)\,\d W_t\eqcomma
\end{align}
initialized at $z_0\sim p_{\rm data}(z_0)$.
This forward process defines conditional densities $p(z\vert z_0,t)$ and marginals $p(z,t)$ for all $t$, and is chosen such that the final distribution $p_{\rm prior}(z_1)$ is a simple prior, typically Gaussian.
Once $f$ and $g$ are fixed, these conditionals are fully determined. For affine forward drifts they are Gaussian in closed form,
\begin{align}
p(z\vert z_0,t) = \mathcal N\big(z;\,\alpha(t)\,z_0,\;\sigma^2(t)\,\mathbb{I}\big)\eqcomma
\label{eq:forward_conditional_affine}
\end{align}
with $\alpha$ and $\sigma$ set by $f$ and $g$, so that we can draw a noised state in a single step without simulating the SDE. This one-shot samplability makes the denoising objective below practical.
The exact generative dynamics are given again by the reverse-time stochastic differential equation,
\begin{align}
\d z
=
\Big(f(z,t)-g^2(t)\,\nabla_z\log p(z,t)\Big)\,\d t
+ g(t)\,\d \bar W_t\eqcomma
\label{eq:reverse_sde_score}
\end{align}
restated here explicitly for the learned setting.
The central difficulty is therefore the estimation of the score $\nabla_z\log p(z,t)$.

Let $s_\theta(z,t)$ be a parameterized approximation of the score, for instance a neural network with tunable parameters $\theta$. 
Replacing $\nabla_z\log p(z,t)$ by $s_\theta(z,t)$ defines an approximate reverse drift, and matching the resulting transition kernel to the true one yields a natural training objective.
Both kernels carry the Gaussian form of \cref{eq:short_time_kernel_rev} with the \emph{same} covariance $g^2(t_k)\,\Delta t\,\mathbb{I}$, so their Kullback--Leibler divergence per step is the squared mean difference in units of that covariance,
\begin{align}
\Delta\mu_k
&\equiv
g^2(t_k)\,\Delta t\,\big[s_\theta(z,t_k)-\nabla_z\log p(z,t_k)\big]\eqcomma
\notag\\
\mathrm{KL}_k
&=
\frac{\|\Delta\mu_k\|^2}{2\,g^2(t_k)\,\Delta t}
=
\tfrac12\,g^2(t_k)\,\Delta t\,
\big\|s_\theta(z,t_k)-\nabla_z\log p(z,t_k)\big\|^2\eqperiod
\label{eq:kl_per_step}
\end{align}
Summing the steps into a Riemann sum and averaging over states yields, up to $t$-dependent constants, the objective
\begin{align}
\mathcal L_{\rm score}(\theta)
\propto
\mathbb E_{t\sim\pi(t)}\,
\mathbb E_{z\sim p(\cdot,t)}
\Big[
    g^2(t)\,
    \big\|s_\theta(z,t)-\nabla_{z}\log p(z,t)\big\|^2
\Big]\eqcomma
\label{eq:score_matching_ideal}
\end{align}
where $\pi(t)$ is a uniform sampling distribution over times $t\in[0,1]$, and the short-time computation in \cref{eq:kl_per_step} fixes the weight $g^2(t)$.
Since the marginal score $\nabla_{z}\log p(z,t)$ is unknown, this loss is rewritten using the forward conditional distribution.
For diffusion processes, the marginal density at time $t$ reads
\begin{align}
p(z,t)
=
\int \d z_0\;
p(z\vert z_0,t)\,p_{\rm data}(z_0)\eqperiod
\label{eq:marginal_from_conditional}
\end{align}
Differentiating with respect to $z$ and using Bayes' rule gives the pointwise identity
\begin{align}
\nabla_{z}\log p(z,t)
=
\mathbb E_{z_0\sim p(\cdot\vert z,t)}
\big[
    \nabla_{z}\log p(z\vert z_0,t)
\big]\eqperiod
\label{eq:score_identity}
\end{align}
Equivalently, once averaged over $z\sim p(\cdot,t)$, this posterior expectation becomes a joint expectation over the forward noising process,
\begin{align}
\mathbb E_{z\sim p(\cdot,t)}
\mathbb E_{z_0\sim p(\cdot\vert z,t)}
\Big[
    F(z_0,z,t)
\Big]
=
\mathbb E_{z_0\sim p_{\rm data}}
\mathbb E_{z\sim p(\cdot\vert z_0,t)}
\Big[
    F(z_0,z,t)
\Big]\eqcomma
\label{eq:posterior_to_forward_joint}
\end{align}
for any suitable test function $F$.
This follows from the two equivalent factorizations of the same joint distribution,
\begin{align}
p(z,t)\,p(z_0\vert z,t)
=
p(z\vert z_0,t)\,p_{\rm data}(z_0)\eqperiod
\end{align}
Thus, we never sample the posterior $p(z_0\vert z,t)$ explicitly. Drawing first $z_0\sim p_{\rm data}$ and then $z\sim p(\cdot\vert z_0,t)$ evaluates the same joint expectation.
Using this relation, we replace the intractable objective from \cref{eq:score_matching_ideal} by the denoising score-matching (DSM) loss
\begin{align}
    \mathcal L_{\rm DSM}(\theta)
    =
    \mathbb E_{t\sim\pi(t)}\,
    \mathbb E_{z_0\sim p_{\rm data}}\,
    \mathbb E_{z\sim p(\cdot\vert z_0,t)}
    \Big[
        g^2(t)\,
        \big\|s_\theta(z,t)
        -\nabla_{z}\log p(z\vert z_0,t)\big\|^2
    \Big]\eqperiod
    \label{eq:dsm_general}
\end{align}
For this choice of weighting, \cref{eq:dsm_general} differs from \cref{eq:score_matching_ideal} only by a $\theta$-independent constant, and therefore has the same optimum~\cite{Vincent2011,Brehmer:2018eca}.
More generally, practical implementations often replace $g^2(t)$ by a positive weighting function $\lambda(t)$,
\begin{align}
    \mathcal L_{\rm DSM}^{(\lambda)}(\theta)
    =
    \mathbb E_{t\sim\pi(t)}\,
    \mathbb E_{z_0\sim p_{\rm data}}\,
    \mathbb E_{z\sim p(\cdot\vert z_0,t)}
    \Big[
        \lambda(t)\,
        \big\|s_\theta(z,t)
        -\nabla_{z}\log p(z\vert z_0,t)\big\|^2
    \Big]\eqcomma
    \label{eq:dsm_weighted}
\end{align}
which changes the relative emphasis placed on different noise levels but not the pointwise target score. 
In the idealized infinite-capacity limit, any strictly positive $\lambda(t)$ therefore has the same score-matching optimum, while for finite networks and finite data the choice of $\lambda(t)$ affects the practical training dynamics and accuracy across time. The choice $\lambda = g^2$ is the \emph{likelihood weighting}, for which the objective bounds the negative log-likelihood~\cite{Song2021mle,huang2021}, while more general weightings admit a variational reading as a weighted integral of evidence lower bounds~\cite{KingmaGao2023}.

After training, generative sampling proceeds by integrating the learned reverse-time SDE obtained from \cref{eq:reverse_sde_score} with the score replaced by $s_\theta$.
Exact likelihood evaluation via the stochastic path integral remains intractable in general. With an exact score, integrating the probability-flow ODE together with a log-density evolution evaluates the model likelihood. With a learned score, the result is the likelihood of the learned deterministic flow.
In the language of \cref{sec:om_action}, score matching is thus the estimation of the drift entering the master kernel. The trained score defines $f^{\theta}_{\mathrm{rev}} = f - g^2 s_\theta$, and inserting it into \cref{eq:kernel_endpoint_def} specifies the path integral of the learned model, which stochastic sampling evaluates trajectory by trajectory.
Discrete-time diffusion models such as denoising diffusion probabilistic models (DDPMs) correspond to specific time discretizations of this construction, combined with particular parameterizations of the score function and the reverse-time dynamics.

\subsubsection*{Conditional flow matching}

Conditional flow matching (CFM) provides a deterministic alternative to stochastic diffusion models that nevertheless avoids direct likelihood evaluation.
In the present framework, CFM replaces stochastic reverse-time sampling by a \emph{probability-flow ordinary differential equation} that preserves the same time-marginal distributions as an underlying diffusion process.
Given a diffusion process with drift $f(z,t)$ and diffusion $g(t)$, the probability-flow ODE follows from rewriting the Fokker--Planck equation~\labelcref{eq:fpe} as a continuity equation.
Using $\nabla_z p = p\,\nabla_z\log p$, we recast the diffusion term as a transport term,
\begin{align}
    \tfrac12 g^2(t)\,\nabla_z^2 p
    = \nabla_z\cdot\big(\tfrac12 g^2(t)\,p\,\nabla_z\log p\big)\eqcomma
    \label{eq:fpe_transport}
\end{align}
so that \cref{eq:fpe} becomes a pure continuity equation,
\begin{align}
\partial_t p(z,t) = -\nabla_z\!\cdot\big(v(z,t)\,p(z,t)\big)\eqcomma
\label{eq:continuity_equation}
\end{align}
with the effective velocity
\begin{align}
\dot z
= v(z,t)
= f(z,t) - \tfrac{1}{2} g^2(t)\,\nabla_z \log p(z,t)\eqperiod
\label{eq:probability_flow_ode}
\end{align}
Because the marginals obey \cref{eq:continuity_equation}, the deterministic flow along \cref{eq:probability_flow_ode} reproduces exactly the same marginals $p(z,t)$ as the stochastic process, with no noise in the sampling dynamics.
The factor $\tfrac12$ originates from rewriting the diffusion term $\tfrac12 g^2\nabla_z^2 p$ of \cref{eq:fpe} as the transport term in \cref{eq:fpe_transport}.
It is therefore \emph{not} the score coefficient of the reverse-time SDE drift, which carries the full $g^2\nabla_z\log p$.
The two descriptions generate identical time-marginals $p(z,t)$ but different trajectory ensembles. The ODE reproduces every single-time marginal of the stochastic process while suppressing its path-space fluctuations.
As in diffusion models, the score $\nabla_z\log p(z,t)$ is unknown and must be approximated.

Rather than estimating the score directly, conditional flow matching constructs tractable training targets by introducing conditional probability flows. Let
\begin{align}
    z = \psi(t;z_0,z_1)\qquad \mwith \qquad z_0\sim p_{\rm data} \qquad \mand \qquad z_1\sim p_{\rm prior}\eqcomma
\end{align}
denote a prescribed interpolation between data samples $z_0$ and latent samples $z_1$.
This defines a \emph{conditional}, \ie endpoint-dependent, target velocity
\begin{align}
v^\star(z,t|z_0,z_1)
=
\partial_t \psi(t;z_0,z_1)\eqcomma
\end{align}
which is known analytically by construction.
We then introduce a neural velocity field $v_\theta(z,t)$ that only depends on the current state and time, not on the particular endpoints $(z_0,z_1)$ used to construct the interpolation. We train it to approximate the \emph{conditional expectation} of this target given $(z,t)$,
\begin{align}
v_\theta(z,t)
\approx
\mathbb E\!\left[
    v^\star(z,t|z_0,z_1)
    \mid z,t
\right]=v(z,t)\eqcomma
\end{align}
leading to the conditional flow matching objective~\cite{Lipman2023}
\begin{align}
\mathcal L_{\rm CFM}(\theta)
=
\mathbb E_{t,z_0,z_1}
\Big[
    \big\|
        v_\theta(z,t)
        - v^\star(z,t| z_0,z_1)
    \big\|^2
\Big]\eqperiod
\label{eq:cfm_loss}
\end{align}
Minimizing this conditional-target loss is equivalent to minimizing the (intractable) marginal-target loss, even though the marginal velocity itself is never evaluated, by the same conditional-versus-marginal replacement that carried the score-matching objective in \cref{eq:score_matching_ideal} to its denoising form in \cref{eq:dsm_general}~\cite{Lipman2023,Lipman2024guide}.

For the Gaussian conditional paths used here, CFM is a deterministic, time-local evaluation principle for the stochastic path integral, sampling along the probability-flow ODE in \cref{eq:probability_flow_ode} while regressing on conditional targets that respect the same marginals. 
For more general interpolants the conditional paths need not derive from a prescribed diffusion, and CFM independently defines transport through the continuity equation.
Even for deterministic CNFs one may prefer CFM over likelihood-based training, since the latter requires evaluating a continuous-time divergence integral that can be computationally expensive in high dimensions.

\subsubsection*{Schrödinger bridges and stochastic control}

Schrödinger bridges (SBs) and stochastic-control approaches provide an alternative evaluation principle~\cite{DeBortoli2023, Shi2023}.
Rather than fixing the forward diffusion and learning the reverse drift via score matching, SB methods determine an \emph{optimal controlled diffusion} whose path measure interpolates between prescribed endpoint marginals.
In the language of the master action this is still nothing but a choice of reverse drift inserted into the same path integral.
It arises globally, as the Doob $h$-transform of the reference dynamics~\cite{Doob1984,Leonard2014,Chen2021sinkhorn}. 
The transform is a reweighting of the reference path measure by a space--time harmonic function $h(z,t)$, which shifts the reference drift by $g^2\,\nabla_z\log h$, a correction of the same form as the score term in \cref{eq:rev_sde}, with $h$ fixed by the two endpoint marginals rather than locally by a prescribed forward process.
Concretely, let $\mathcal P_0[z]$ be a reference path measure induced by a known diffusion, \eg Brownian motion or an Ornstein--Uhlenbeck process.
Among all path measures $\mathcal P[z]$ with the prescribed endpoint marginals, the Schrödinger bridge solves
\begin{align}
\mathcal P^\star
=
\arg\min_{\mathcal P}
\Big\{
    \mathrm{KL}\!\big(\mathcal P \,\|\, \mathcal P_0\big)
    \;\big|\;
    \mathcal P(z(0))=p_{\rm data},\;
    \mathcal P(z(1))=p_{\rm prior}
\Big\}\eqperiod
\label{eq:schrodinger_bridge}
\end{align}
This is an entropy-regularized optimal-transport problem on path space~\cite{Leonard2014}, in which the SB is the path measure closest to the reference while transporting $p_{\rm data}$ to $p_{\rm prior}$.

In the path-integral language, \cref{eq:schrodinger_bridge} is a variational principle on the Onsager--Machlup path measure.
Writing the reference measure as $\mathcal P_0\propto e^{-S_0[z]}$, with $S_0$ its Onsager--Machlup action, the KL divergence is
\begin{align}
    \mathrm{KL}(\mathcal P\,\|\,\mathcal P_0)
    = \mathbb{E}_{\mathcal P}\big[\,S_0[z]\,\big]
      - \mathbb{H}[\mathcal P] + \mathrm{const}\eqcomma
\end{align}
where $\mathbb{H}[\mathcal P]$ is the path entropy.
The SB is thus the entropy-regularized minimum of the Onsager--Machlup action over path measures with the prescribed boundary marginals.
In contrast to score-based diffusion, which fixes the forward process and matches the reverse drift \emph{locally}, the SB determines the entire probability flow \emph{globally}, subject to both endpoint constraints.

\emph{Iterative proportional fitting} (IPF) solves the constrained problem in \cref{eq:schrodinger_bridge}~\cite{DeBortoli2023}. 
Starting from $\mathcal P^{(0)}=\mathcal P_0$, one alternates two half-bridge KL projections,
\begin{alignat}{3}
\mathcal P^{(2n+1)}
&= \arg\min_{\mathcal P}\;
   \mathrm{KL}\!\big(\mathcal P \,\|\, \mathcal P^{(2n)}\big)
   &\qquad \mwith\qquad
   \mathcal P(z(1)) &= p_{\rm prior}\eqcomma
\notag\\
\mathcal P^{(2n+2)}
&= \arg\min_{\mathcal P}\;
   \mathrm{KL}\!\big(\mathcal P \,\|\, \mathcal P^{(2n+1)}\big)
   &\qquad \mwith \qquad
   \mathcal P(z(0)) &= p_{\rm data}\eqcomma
\label{eq:ipf_iteration}
\end{alignat}
each step enforcing one endpoint marginal while staying as close as possible to the current iteration.
The iteration converges to $\mathcal P^\star$~\cite{DeBortoli2023}.
In practice each half-bridge is itself a diffusion. 
Its optimal drift is a score, and a score-matching regression of the form of \cref{eq:dsm_general} solves it, so that one IPF step reduces to training a diffusion model against the trajectories of the previous iteration~\cite{DeBortoli2023}.
Subsequent formulations recast the same alternation as iterative Markovian fitting~\cite{Shi2023}, which projects alternately onto the Markov and reciprocal classes of path measures.

\subsubsection{Variational autoencoders as variational approximations}
\label{sec:vae_as_variational}

Variational autoencoders (VAEs) operate directly at the level of the latent-variable model, without introducing the temporal chain in \cref{eq:lvm_chain}.
In the language of this paper they correspond not to a particular dynamics but to the absence of one, \ie a single latent variable $z$ with prior $p_{\rm prior}(z)$ and decoder $p_\theta(x\vert z)$,
\begin{align}
    p_\theta(x) = \int \d z\; p_\theta(x\vert z)\,p_{\rm prior}(z)\eqcomma
    \label{eq:vae_marginal}
\end{align}
and no path over intermediate states.
The marginal likelihood is intractable for expressive decoders, since the integral over $z$ cannot be computed in closed form.
Thus, the true posterior
\begin{align}
    p_\theta(z\vert x)
    = \frac{p_\theta(x\vert z)\,p_{\rm prior}(z)}{p_\theta(x)}\eqcomma
    \label{eq:vae_true_posterior}
\end{align}
is likewise intractable.
Introducing an approximate encoder $q_\phi(z\vert x)$ and writing
\begin{align}
    \log p_\theta(x)
    &= \log \int \d z\;
       q_\phi(z\vert x)\,
       \frac{p_\theta(x\vert z)\,p_{\rm prior}(z)}{q_\phi(z\vert x)}\eqcomma
\end{align}
Jensen's inequality applied to the concave logarithm yields
\begin{align}
    \log p_\theta(x)
    &\geq
    \mathbb{E}_{q_\phi(\cdot\vert x)}\!\left[\log p_\theta(x\vert z)\right]
    - \mathrm{KL}\big(q_\phi(z\vert x)\,\|\,p_{\rm prior}(z)\big)
    \equiv \mathcal{L}_{\rm ELBO}(x;\theta,\phi)\eqperiod
    \label{eq:elbo}
\end{align}
The gap between the ELBO and the true log-likelihood is exactly the
KL divergence between the approximate and true posteriors,
\begin{align}
    \log p_\theta(x) - \mathcal{L}_{\rm ELBO}
    = \mathrm{KL}\big(q_\phi(z\vert x)\,\|\,p_\theta(z\vert x)\big)
    \geq 0\eqcomma
    \label{eq:elbo_gap}
\end{align}
which follows directly from the definition of KL divergence.
Maximizing the ELBO simultaneously tightens this bound and minimizes the KL between $q_\phi$ and the true posterior. 
The bound is tight when $q_\phi = p_\theta(\cdot\vert x)$.
Both conditionals are distributions even though deterministic networks implement them. 
Each network returns the parameters of a Gaussian, the decoder the mean of an observation model and the encoder a mean and a variance,
\begin{align}
    p_\theta(x\vert z) = \mathcal{N}\big(x;\,\mu_\theta(z),\,\sigma^2\,\mathbb{I}\big)
    \qquad \mand \qquad
    q_\phi(z\vert x) = \mathcal{N}\big(z;\,\mu_\phi(x),\,\sigma_\phi^2(x)\,\mathbb{I}\big)\eqcomma
    \label{eq:vae_decoder}
\end{align}
so the reconstruction term of \cref{eq:elbo} equals the familiar squared-error loss with $\sigma^2$ setting its weight against the KL term.
Neither spread is decorative, and the two fail at opposite ends of \cref{eq:elbo}. A deterministic decoder means $\sigma\to0$, where $p_\theta(x\vert z)\to\delta(x-\mu_\theta(z))$ and the reconstruction term diverges, while a deterministic encoder gives $\mathrm{KL}(\delta\,\|\,p_{\rm prior})=\infty$ and drives the bound to $-\infty$.
The decoder noise makes the model a density, while the encoder noise lets $q_\phi$ approximate a posterior that has support.

One may nonetheless view the single-latent model as the degenerate endpoint of the path construction, in which the trajectory consists of a single transition of unit duration. 
This analogy is formal rather than dynamical, since a VAE does not arise from a continuum limit, and we include it only for completeness.
The single transition kernel $p(z_{0}|z_1)$ plays the role of the short-time kernel in \cref{eq:short_time_kernel_rev} integrated over a single step of unit duration.
The discrete action is
\begin{align}
    S_{\theta}[x, z]
    = -\log p_\theta(x\vert z)\eqcomma
\end{align}
so the path integral reduces to a single ordinary integral over the latent variable,
\begin{align}
    p_\theta(x)
    = \int \d z\; p_{\rm prior}(z)\,e^{-S_{\theta}[x, z]}\eqcomma
\end{align}
which is exactly the latent-variable model above.
The ELBO is then a variational lower bound obtained by importance-weighting this integral with $q_\phi$,
\begin{align}
    \log p_\theta(x)
    &= \log \mathbb{E}_{q_\phi(\cdot\vert x)}\!\left[
        \frac{p_{\rm prior}(z)\,e^{-S_{\theta}[x, z]}}{q_\phi(z\vert x)}
      \right]
    \geq
    \mathbb{E}_{q_\phi(\cdot\vert x)}\!\left[
        -S_{\theta}[x, z] - \log\frac{q_\phi(z\vert x)}{p_{\rm prior}(z)}
    \right]\eqperiod
\end{align}
In the path-integral language, maximizing the ELBO is equivalent to finding the importance distribution $q_\phi$ that best approximates the posterior path measure $p_\theta(z\vert x)$ within the parametric family $\{q_\phi\}$.

Within the present unifying perspective, VAEs are thus characterized by two approximations:
(i) the use of a single latent variable rather than a latent trajectory, and (ii) a variational approximation, replacing the intractable posterior by the tractable family $q_\phi(z\vert x)$.
The tightness of the bound is controlled by the expressiveness of $q_\phi$. When $q_\phi = p_\theta(\cdot\vert x)$, the gap in \cref{eq:elbo_gap} vanishes and the ELBO recovers the true log-likelihood.

\subsubsection*{Adversarial models as likelihood-free saddle points}

Setting $\sigma=0$ in \cref{eq:vae_decoder} makes the decoder deterministic, $p_\theta(x\vert z)\to\delta(x-\mu_\theta(z))$, in the same way that $g\to0$ collapses the path integral onto its saddle in \cref{sec:nf_saddle}. The marginal in \cref{eq:vae_marginal} then reads
\begin{align}
    p_\theta(x)
    =
    \int \d z\; \delta\big(x-\mu_\theta(z)\big)\,p_{\rm prior}(z)\eqperiod
    \label{eq:gan_pushforward}
\end{align}
Whether $p_\theta$ still has a density depends on $\mu_\theta$ alone, and five cases arise.
\begin{enumerate}[label=(\roman*),ref=(\roman*)]
    \item\label{case:vae} At $\sigma>0$ the decoder is a Gaussian of finite width, so each $\mu_\theta(z)$ contributes to $p_\theta(x)$ at every $x$ and the marginal stays positive everywhere. The decoder width alone guarantees the density, whatever the latent dimension, and we remain in the variational setting above.

    \item\label{case:flow} At $\sigma=0$ with an invertible $\mu_\theta$ between spaces of equal dimension, \cref{eq:change_of_variable_delta} rewrites the delta and \cref{eq:gan_pushforward} returns the change of variables, recovering the normalizing flow of \cref{sec:nf_saddle}.

    \item\label{case:branches} At $\sigma=0$ with equal dimensions but no inverse, \cref{eq:change_of_variable_delta} generalizes to a sum over the solutions of $\mu_\theta(z)=x$,
    \begin{align}
        \delta\big(x-\mu_\theta(z)\big)
        =
        \sum_{y\,\in\,\mu_\theta^{-1}(x)}
        \frac{\delta(z-y)}
             {\left|\det \dfrac{\partial \mu_\theta(z)}{\partial z}\right|_{z=y}}\eqcomma
        \label{eq:gan_delta_multivalued}
    \end{align}
    where $\mu_\theta^{-1}(x)$ is now a discrete set rather than a single point. Inserting this into \cref{eq:gan_pushforward} gives
    \begin{align}
        p_\theta(x)
        =
        \sum_{y\,\in\,\mu_\theta^{-1}(x)}
        \frac{p_{\rm prior}(y)}
             {\left|\det \dfrac{\partial \mu_\theta(z)}{\partial z}\right|_{z=y}}\eqcomma
        \label{eq:gan_multivalued}
    \end{align}
    the multibranch form of \cref{eq:nf_direct}. A one-dimensional example makes this concrete. The map $\mu_\theta(z)=z^2$ reaches only $x\geq0$, so $p_\theta$ vanishes for $x<0$, while every $x>0$ has the two solutions $y=\pm\sqrt{x}$ with derivative of modulus $2\sqrt{x}$, and \cref{eq:gan_multivalued} gives
    \begin{align}
        p_\theta(x)
        =
        \frac{p_{\rm prior}(\sqrt{x})+p_{\rm prior}(-\sqrt{x})}{2\sqrt{x}}
        \qquad \mwith \qquad x>0\eqperiod
        \label{eq:gan_square_example}
    \end{align}
    The density is exact and normalized although the map has no inverse, and it diverges at $x\to0$, the image of the critical point $z=0$ where the derivative vanishes.
    Enumerating the solutions of $\mu_\theta(z)=x$ for a generic network is a global root-finding problem with no guarantee of completeness, which renders $p_\theta(x)$ intractable.

    \item\label{case:fiber} At $\sigma=0$ with a latent space larger than the data space, the solutions of $\mu_\theta(z)=x$ form a surface of dimension $\dim z-\dim x$ rather than a discrete set. 
    The sum in \cref{eq:gan_multivalued} becomes an integral over that surface with its induced measure and Jacobian factor, which again exists but is generally intractable.

    \item\label{case:singular} At $\sigma=0$ with a latent space smaller than the data space, or with a Jacobian that loses rank on an open set, $\mu_\theta$ maps into a subset of the data space of zero volume. The distribution of generated samples remains well defined as a measure, but it assigns all its weight to that subset, so no density $p_\theta(x)$ exists and \cref{eq:elbo} has no log-likelihood left to bound. Samples can still be drawn, but $p_\theta$ cannot be evaluated at any point.
\end{enumerate}
Invertibility is therefore sufficient for a tractable likelihood but not necessary, and a bottleneck is not by itself the obstruction, as VAEs use one freely. Only at $\sigma=0$ does the geometry of $\mu_\theta$ decide the question, and in cases \labelcref{case:branches,case:fiber,case:singular} it leaves us without a likelihood we can evaluate. Training then has to proceed from samples alone.

Generative adversarial networks do this~\cite{Goodfellow2014}. A classifier $D$ and the map $\mu_\theta$ share a single objective,
\begin{align}
    \mathcal{L}_{\rm GAN}(D,\theta)
    =
    \mathbb{E}_{x\sim p_{\rm data}}\big[\log D(x)\big]
    +
    \mathbb{E}_{x\sim p_\theta}\big[\log\big(1-D(x)\big)\big]\eqcomma
    \label{eq:gan_discriminator_loss}
\end{align}
which the classifier maximizes and the generator minimizes. When both densities exist, maximizing pointwise in $D(x)$ at fixed $\theta$ gives
\begin{align}
    D^\star(x) = \frac{p_{\rm data}(x)}{p_{\rm data}(x)+p_\theta(x)}
    \qquad \mand \qquad
    \log\frac{D^\star(x)}{1-D^\star(x)} = \log\frac{p_{\rm data}(x)}{p_\theta(x)}\eqperiod
    \label{eq:gan_optimal_discriminator}
\end{align}
The optimal classifier returns the likelihood ratio that the Neyman--Pearson lemma identifies as the most powerful test statistic between two simple hypotheses~\cite{NeymanPearson1933}.
Inserting $D^\star$ back into \cref{eq:gan_discriminator_loss} leaves the generator minimizing
\begin{align}
    \mathcal{L}_{\rm GAN}(D^\star,\theta)
    =
    2\,\mathrm{JS}\big(p_{\rm data}\,\|\,p_\theta\big) - \log 4
    \eqcomma
    \label{eq:gan_js}
\end{align}
where $\mathrm{JS}$ is the Jensen--Shannon divergence, which vanishes when $p_\theta=p_{\rm data}$. The generator improves $p_\theta$ without ever evaluating it, since the density enters only through the ratio that the classifier estimates from samples.

All three put the likelihood out of reach, but only case~\labelcref{case:singular} breaks \cref{eq:gan_optimal_discriminator}. When $p_\theta$ is a density but we merely cannot evaluate it, the ratio is well defined and the classifier estimates it from samples exactly as intended. When $p_\theta$ is singular, its support and that of $p_{\rm data}$ generically fail to overlap, which a bottleneck generator enforces by construction and natural data is believed to do on its own, and no optimal discriminator as in \cref{eq:gan_optimal_discriminator} exists.
A classifier that separates the two supports perfectly then maximizes \cref{eq:gan_discriminator_loss}, the Jensen--Shannon divergence equals $\log2$ for every such pair, and \cref{eq:gan_js} stays constant in $\theta$ with vanishing gradient~\cite{Arjovsky2017}. Smearing both distributions with noise before comparing them restores the overlap and with it a usable gradient, so the width removed in \cref{eq:gan_pushforward} returns in the training criterion rather than in the model.
The smearing need not be applied to the samples. Expanding the discriminator to leading order in the noise variance turns the convolution into a penalty on its gradient norm~\cite{Roth2017}, so a gradient-penalized discriminator minimizes the smoothed divergence while the objective stays untouched, and such penalties restore local convergence even when both distributions lie on lower-dimensional manifolds~\cite{Mescheder2018}. 
Constraining the discriminator through spectral normalization~\cite{Miyato2018} limits how sharply it can separate the two supports and acts in the same direction.

\begin{table}[t]
\centering
\begin{small}
\begin{tabular}{l c c l l l}
\toprule
Model
& Forward
& Reverse
& Loss
& Likelihood
& Path-integral picture \\
\midrule
NF/CNF
& Det.
& Det.
& Likelihood
& Exact
& Saddle point ($g\to0$) \\

CFM
& Cond.\ path
& Det.
& Cond.\ flow matching
& ODE-based
& Probability-flow ODE \\

DDPM/SM
& Stoch.
& Stoch.
& Denoising/score matching
& Intractable
& Full path measure \\

SB
& Stoch.
& Stoch.
& Path-space KL (IPF)
& Intractable
& Glob.\ path-space var. \\

VAE
& Stoch.
& Stoch.
& ELBO
& Lower bound
& Single-latent var. \\
GAN
& --
& Det.
& Adversarial
& None/Intractable
& Saddle point ($g\to0$) \\
\bottomrule
\end{tabular}
\end{small}
\caption{
Comparison of generative models within the common framework.
``Forward'' and ``Reverse'' denote the dynamics used to propagate probability mass from data to latent space and back, while ``Det.'' and ``Stoch.'' indicate deterministic and stochastic sampling dynamics, and ``Cond.\ path'' a prescribed conditional interpolation.
``ODE-based'' likelihood means the density is available only by integrating the probability-flow ODE, not in closed form.
The final column gives the role each model plays within the path-integral formulation of \cref{sec:path_integral}.
}
\label{tab:model_taxonomy}
\end{table}

\subsubsection{Summary: evaluation principles for the master path integral}

All generative models discussed in this section are organized by the same path integral in \cref{eq:path_integral_OM}. 
Their differences arise not from distinct probabilistic foundations but from how that path integral is evaluated or approximated in practice, while variational autoencoders and adversarial models arise as degenerate single-step and likelihood-free realizations of the underlying latent-variable formulation.
\Cref{tab:model_taxonomy} summarizes the resulting taxonomy.
Viewed through this unified lens, the apparent diversity of modern generative models reflects different evaluation strategies within a common latent-variable construction, and for the continuous-time transport models these strategies act on the same stochastic path integral.

\clearpage
\section{Probability flows as interacting field theories}
\label{sec:qft}

In \cref{sec:path_integral}, we wrote a wide class of generative models as a continuous probabilistic path integral, with the data density obtained by summing over latent trajectories connecting the prior at $t=1$ to the observation at $t=0$. 
Reverse-time dynamics with drift $f_{\mathrm{rev}}(z,t)$ and
diffusion $g(t)$ lead to the Onsager--Machlup action,
\begin{align}
S_\text{OM}[z]
=\int_0^1 \d t\;
\frac{\|\dot z - f_{\mathrm{rev}}(z,t)\|^2}{2g^2(t)}\eqperiod
\label{eq:qft_om_recall}
\end{align}
Linear (or linearized) probability flows define Gaussian path measures and play the role of free theories, while nonlinearities in the drift generate
interactions. This perspective leads to a diagrammatic expansion and a scattering interpretation of generative probability transport, which we
develop below.

\subsection{Free--interacting split and MSRJD representation}
\label{sec:free_vs_interacting}

The central object in the path-integral formulation is the action $S_\text{OM}[z]$, which determines the path measure and hence the generative model.
From a field-theoretic point of view, the distinction between \emph{free} and \emph{interacting} probability flows is governed by whether the action is quadratic in the path variables.
To make this explicit, we split the reverse drift into a linear (affine) part and a nonlinear remainder,
\begin{align}
f_{\mathrm{rev}}(z,t)
=
A(t)\,z + b(t) + f_{\mathrm{int}}(z,t)\eqcomma
\qquad
f_{\mathrm{int}}(\,\cdot\,,t)\;\text{nonlinear in }z\eqperiod
\label{eq:drift_split}
\end{align}
To build intuition, take the usual affine forward process transporting the data to a Gaussian prior. If the data is also an isotropic Gaussian, every forward marginal $p(z,t)$ stays Gaussian, the exact score $\nabla_z\log p$ is linear in $z$, and $f_{\mathrm{rev}}$ is affine, so $f_{\mathrm{int}}=0$ and the theory is free. Non-Gaussian data, such as multimodal targets, make the intermediate scores nonlinear and generate $f_{\mathrm{int}}$, which therefore measures how far the transport departs from connecting two Gaussians.
Inserting \cref{eq:drift_split} into the Onsager--Machlup action yields
\begin{align}
S_\text{OM}[z]
=
\int_0^1 \d t\;
\frac{\|\dot z - A(t)\,z - b(t)\|^2}{2g^2(t)}
\;+\;
S_{\mathrm{int}}[z]\eqcomma
\label{eq:action_split}
\end{align}
where the interaction functional is defined by
\begin{align}
S_{\mathrm{int}}[z]
\equiv
\int_0^1 \d t\;
\frac{1}{2g^2(t)}
\Big(
\|f_{\mathrm{int}}(z,t)\|^2
-2\,(\dot z - A(t)\,z - b(t))\cdot f_{\mathrm{int}}(z,t)
\Big)\eqperiod
\label{eq:om_interaction_functional}
\end{align}
If $f_{\mathrm{int}}=0$, the action is quadratic in $z$ and the path integral is Gaussian, so all correlation functions are determined by the corresponding propagators.
Conversely, any nonlinear dependence of the drift generates non-quadratic terms in $S_{\mathrm{int}}$, which act as interaction vertices in a perturbative expansion.

While \cref{eq:om_interaction_functional} defines interactions directly at the level of the Onsager--Machlup action, it is not always the most convenient representation for perturbation theory.
In particular, the Onsager--Machlup form is second order in time derivatives and entangles drift and noise in a way that obscures the causal structure of the underlying dynamics. 
For diagrammatic calculations it is therefore advantageous to switch to a first-order functional integral representation, in which interactions enter linearly and the response structure is explicit. 

The MSRJD~\cite{Martin:1973zz, Janssen:1976qag, dominicis1976techniques} construction provides exactly this representation, which we derive by considering again the reverse-time dynamics
\begin{align}
\d z = f_{\mathrm{rev}}(z,t)\,\d t + g(t)\,\d \bar W_t\eqcomma
\label{eq:msrjd_sde_start}
\end{align}
and rewrite the associated path weight as a first-order functional integral over $z$ and an auxiliary field.
Our central object is the path weight from \cref{eq:om_weight},
\begin{align}
e^{-S_\text{OM}[z]} = \exp\!\left[-\int_0^1 \d t\;\frac{\|\dot z - f_{\mathrm{rev}}(z,t)\|^2}{2g^2(t)}\right]\eqcomma
\label{eq:msrjd_om_weight}
\end{align}
which we bring to MSRJD form.
First, we represent this Gaussian as a noise average of a functional delta enforcing the equation of motion.
Introducing the reverse white-noise field $\xi(t)$, the continuum limit of the discrete noise $\xi_k$ in \cref{eq:em_reverse_step}, the dynamics take the Langevin form
\begin{align}
\dot z(t) = f_{\mathrm{rev}}(z(t),t) + g(t)\,\xi(t)
\qquad \mwith \qquad
\big\langle\xi(t)\,\xi(t')^{\!\top}\big\rangle=\delta(t-t')\,\mathbb{I}\eqcomma
\label{eq:msrjd_langevin}
\end{align}
and introducing the delta enforcing \cref{eq:msrjd_langevin} gives
\begin{align}
e^{-S_\text{OM}[z]}
=
\int \mathcal D\xi\;
\exp\!\left[-\frac12\int_0^1\d t\;\|\xi(t)\|^2\right]
\delta\!\left[\dot z - f_{\mathrm{rev}}(z,t) - g(t)\,\xi(t)\right]\eqcomma
\label{eq:msrjd_noise_average}
\end{align}
since the $\xi$-integral against the delta sets $g\xi = \dot z - f_{\mathrm{rev}}$ and returns \cref{eq:msrjd_om_weight}.
Second, we resolve the delta by its functional Fourier representation~\cite{Zinn-Justin:2002ecy,Tauber2014,Kamenev2011}, which introduces the real \emph{response field} $\hat z(t)$,
\begin{align}
\delta\!\left[\dot z - f_{\mathrm{rev}} - g\xi\right]
\propto
\int \mathcal D \hat z\;
\exp\!\left[
-\imag\int_0^1 \d t\;
\hat z(t)\cdot
\Big(\dot z(t)-f_{\mathrm{rev}}(z(t),t)-g(t)\,\xi(t)\Big)
\right]\eqperiod
\label{eq:msrjd_delta_fourier}
\end{align}
Third, the noise field $\xi(t)$ now appears linearly and is integrated out by completing the square, a Hubbard--Stratonovich transformation~\cite{Stratonovich:1957SPhD,Hubbard:1959ub} that trades the linear noise coupling for the quadratic response term,
\begin{align}
\int \mathcal D\xi\;
\exp\!\left[
-\int_0^1\d t\,
\Big(\frac12\,\|\xi\|^2 + \imag\,g(t)\,\hat z\cdot\xi\Big)
\right]
=
\exp\!\left[
-\frac12\int_0^1\d t\;
g^2(t)\,\|\hat z\|^2
\right]\eqperiod
\label{eq:msrjd_noise_integral}
\end{align}
Collecting the three steps expresses the weight as a functional integral over the response field,
\begin{align}
e^{-S_\text{OM}[z]}
=
\int\mathcal D\hat z\;e^{-S_{\mathrm{MSRJD}}[z,\hat z]}\eqcomma
\label{eq:msrjd_weight_identity}
\end{align}
with the measure $\mathcal D\hat z$ absorbing the $z$-independent normalization of the Fourier representation, as $\mathcal D z$ absorbs that of the Onsager--Machlup weight in \cref{sec:om_action}, and the MSRJD action
\begin{align}
S_{\mathrm{MSRJD}}[z,\hat z]
=
\int_0^1 \d t\;
\left[
\imag\,\hat z\cdot\big(\dot z - f_{\mathrm{rev}}(z,t)\big)
+\frac12\,g^2(t)\,\|\hat z\|^2
\right]\eqcomma
\label{eq:msrjd_action}
\end{align}
where we use the It\^o convention adopted in \cref{sec:om_action}. With causal time ordering, the functional determinant from the delta constraint is triangular with a field-independent diagonal, and we absorb it into the normalization of $\mathcal D\hat z$. Other discretizations generate an additional local divergence term and would modify the interaction vertices accordingly.

The MSRJD representation provides a first-order functional integral in which nonlinearities in the drift enter linearly through the coupling $\imag\,\hat z\cdot f_{\mathrm{rev}}(z,t)$. 
For real $\hat z$ the noise term damps the integrand, so the representation is convergent as it stands. Equivalently, the factor of $\imag$ can be absorbed into a response field integrated along the imaginary axis, which renders the action real~\cite{Zinn-Justin:2002ecy, Tauber2014}.
Using the decomposition from \cref{eq:drift_split}, the action becomes
\begin{align}
S_{\mathrm{MSRJD}}[z,\hat z]
=
S_0[z,\hat z]
+ S_{\mathrm{int}}[z,\hat z]\eqcomma
\label{eq:msrjd_split}
\end{align}
with the free part
\begin{align}
S_0[z,\hat z]
=
\int_0^1 \d t\;
\left[
\imag\,\hat z\cdot\big(\dot z - A(t)\,z - b(t)\big)
+\frac12\,g^2(t)\,\|\hat z\|^2
\right]\eqcomma
\label{eq:msrjd_free}
\end{align}
and the interaction
\begin{align}
S_{\mathrm{int}}[z,\hat z]
=
-\,\imag\int_0^1 \d t\;
\hat z(t)\cdot f_{\mathrm{int}}(z(t),t)\eqperiod
\label{eq:msrjd_interaction}
\end{align}
%

\subsection{Scattering of probability flows}
\label{sec:scattering}

The MSRJD representation introduced above allows us to reinterpret generative probability transport as a scattering process in latent space.
In this picture, probability mass is propagated from a simple prior distribution at $t=1$ to the data manifold at $t=0$ by an interacting dynamical evolution, in close analogy with time-dependent scattering in quantum field theory.

The evolution between the two ends is carried by the transition kernel $K(z_0,0\mid z_1,1)$ constructed in \cref{sec:om_action}.
Through the marginal equation \labelcref{eq:scattering_density}, the prior $p_{\rm prior}(z_1)$ plays the role of an \emph{in-state}, the data density $p_{\rm data}(z_0)$ defines an \emph{out-state}, and the kernel encodes the dynamical evolution between them, in analogy with a transition amplitude.

In the absence of interactions, \ie a linear reverse drift and hence a quadratic action, we can evaluate the kernel $K$ exactly.
The resulting Gaussian kernel describes a free probability flow, in which probability mass propagates without mode coupling or distortion beyond what the linear drift and the diffusion induce.
This situation is directly analogous to free-particle propagation, where the transition amplitude is fully determined by the free propagator.

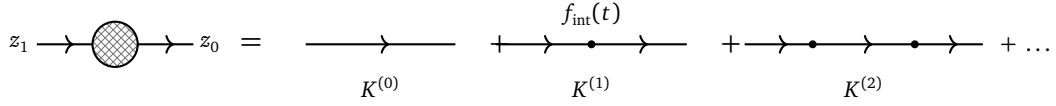
\begin{figure}[t]
\centering
\begin{tikzpicture}[line width=0.9pt, scale=0.9]
  \node[blob, minimum size=17pt] (K) at (0,0) {};
  \draw[fermion] (-1.15,0) -- (K);
  \draw[fermion] (K) -- (1.15,0);
  \node at (-1.4,0) {\footnotesize $z_1$};
  \node at (1.4,0)  {\footnotesize $z_0$};
  \node at (2.0,0) {$=$};
  \begin{scope}[shift={(3.9,0)}]
    \draw[fermion] (-1.1,0) -- (1.1,0);
    \node at (0,-0.62) {\footnotesize $K^{(0)}$};
    \node at (1.75,0) {$+$};
  \end{scope}
  \begin{scope}[shift={(7.0,0)}]
    \draw[fermion] (-1.4,0) -- (0,0);
    \node[vertex] (v) at (0,0) {};
    \draw[fermion] (0,0) -- (1.4,0);
    \node at (0,0.45) {\footnotesize $f_{\mathrm{int}}(t)$};
    \node at (0,-0.62) {\footnotesize $K^{(1)}$};
    \node at (2.05,0) {$+$};
  \end{scope}
  \begin{scope}[shift={(11.0,0)}]
    \draw[fermion] (-1.75,0) -- (-0.75,0);
    \node[vertex] (w1) at (-0.75,0) {};
    \draw[fermion] (-0.75,0) -- (0.75,0);
    \node[vertex] (w2) at (0.75,0) {};
    \draw[fermion] (0.75,0) -- (1.75,0);
    \node at (0,-0.62) {\footnotesize $K^{(2)}$};
    \node at (2.4,0) {\footnotesize $+\;\dots$};
  \end{scope}
\end{tikzpicture}
\caption{%
Diagrammatic form of the kernel expansion in \cref{eq:kernel_perturbative_expansion_full}. The hatched blob is the full kernel $K(z_0,0\mid z_1,1)$ and the plain arrowed line is free propagation $K^{(0)}$ under the linearized drift. Each dot is one insertion of $f_{\mathrm{int}}$ integrated over its time argument as in \cref{eq:kernel_duhamel}, so $K^{(n)}$ carries $n$ insertions. No form of $f_{\mathrm{int}}$ is assumed. Resolving a dot into vertices with a definite number of legs requires expanding $f_{\mathrm{int}}$ in powers of $z$, which is done in \cref{sec:diagrammatics}. Arrows run along the generative direction, from the prior at $t=1$ to the data at $t=0$.
}
\label{fig:kernel_series}
\end{figure}

Nonlinear probability flows correspond to interacting theories, for which the kernel $K$ cannot be evaluated in closed form.
Rewriting the Onsager--Machlup weight in \cref{eq:kernel_endpoint_def} with the MSRJD identity in \cref{eq:msrjd_weight_identity}, the kernel becomes a two-field functional integral,
\begin{align}
K(z_0,0\mid z_1,1)
=
\int \mathcal D z\,\mathcal D\hat z\;
e^{-S_{\mathrm{MSRJD}}[z,\hat z]}\eqcomma
\label{eq:kernel_msrjd}
\end{align}
with the latent-path boundary values fixed by the arguments of the kernel as in \cref{sec:om_action}, and the response field unconstrained at both ends.
For perturbation theory it is convenient to release the data endpoint and work in the ensemble of the sampler, \ie trajectories launched from the prior point $z_1$ with a free final state.
We therefore define the normalized free expectation value
\begin{align}
\big\langle O[z,\hat z] \big\rangle_{0}
\equiv
\int \d z_0
\int \mathcal D z\,\mathcal D \hat z\;
O[z,\hat z]\,
e^{-S_0[z,\hat z]}
\qquad \mwith \qquad 
\langle 1\rangle_0 = 1\eqcomma
\label{eq:free_average_onesided}
\end{align}
where the inner path integral is pinned at $z(0)=z_0$ and $z(1)=z_1$ as in \cref{eq:kernel_endpoint_def} and the data endpoint is subsequently integrated over.
At $S_{\mathrm{int}}=0$ the inner integral is the transition density of the affine process, normalized as in \cref{eq:kernel_normalization}.
The average describes realizations of the free reverse process started at $z_1$.
Pinning and releasing the data endpoint are related exactly, since inserting a delta function of the endpoint collapses the $z_0$-integral back onto the pinned path integral,
\begin{align}
\big\langle \delta\big(z(0)-z_0\big)\, O[z,\hat z] \big\rangle_{0}
=
\int \mathcal D z\,\mathcal D \hat z\;
O[z,\hat z]\,
e^{-S_0[z,\hat z]}\eqcomma
\label{eq:kernel_delta_insertion}
\end{align}
with the boundary values on the right-hand side again fixed at $z(0)=z_0$ and $z(1)=z_1$.
Splitting $S_{\mathrm{MSRJD}}=S_0+S_{\mathrm{int}}$ into its free and interacting parts, cf.\ \cref{eq:msrjd_split}, and expanding the interaction exponential inside \cref{eq:kernel_msrjd}, each order of the transition kernel becomes a one-sided free average with an endpoint insertion,
\begin{align}
K(z_0,0\mid z_1,1)
&=
\big\langle
\delta\big(z(0)-z_0\big)\,
e^{-S_{\mathrm{int}}[z,\hat z]}
\big\rangle_{0}
\nonumber\\
&=
\sum_{n=0}^{\infty}\frac{(-1)^n}{n!}
\big\langle
\delta\big(z(0)-z_0\big)\,
\big(S_{\mathrm{int}}[z,\hat z]\big)^n
\big\rangle_{0}
\equiv \sum_{n=0}^{\infty} K^{(n)}(z_0,0\mid z_1,1)\eqperiod
\label{eq:kernel_perturbative_expansion_full}
\end{align}
The zeroth order $K^{(0)}=\langle\delta(z(0)-z_0)\rangle_0$ is the free kernel, evaluated in closed form in \cref{sec:diagrammatics}, and the first corrections read
\begin{align}
K^{(1)}(z_0,0\mid z_1,1)
&=
-\,\big\langle \delta\big(z(0)-z_0\big)\,S_{\mathrm{int}}[z,\hat z] \big\rangle_{0}\eqcomma
\notag
\\[0.5ex]
K^{(2)}(z_0,0\mid z_1,1)
&=
\frac12\,
\big\langle \delta\big(z(0)-z_0\big)\,\big(S_{\mathrm{int}}[z,\hat z]\big)^2 \big\rangle_{0}\eqperiod
\label{eq:K1_K2_def}
\end{align}
Using the explicit form of \cref{eq:msrjd_interaction}, these become
\begin{align}
K^{(1)}(z_0,0\mid z_1,1)
&=
\imag\int_0^1 \d t\;
\Big\langle
\delta\big(z(0)-z_0\big)\,
\hat z(t)\cdot f_{\mathrm{int}}(z,t)
\Big\rangle_{0}\eqcomma
\notag
\\[0.5ex]
K^{(2)}(z_0,0\mid z_1,1)
&=
-\frac12\int_0^1 \d t_1\,\d t_2\;
\Big\langle
\delta\big(z(0)-z_0\big)\,
\hat z(t_1)\cdot f_{\mathrm{int}}(z,t_1)\;
\hat z(t_2)\cdot f_{\mathrm{int}}(z,t_2)
\Big\rangle_{0}\eqperiod
\label{eq:K1_K2_explicit}
\end{align}
Integrating \cref{eq:kernel_perturbative_expansion_full} over the data endpoint removes the insertion, 
\begin{align}
    \int\d z_0\,K^{(n)} = \frac{(-1)^n}{n!}\langle S_{\mathrm{int}}^n\rangle_0\eqcomma
\end{align}
and every such average vanishes for $n\geq1$. Each response field must contract with a latent field at one of the vertices, so every Wick pattern contains a closed cycle of contractions $\langle z(t)\,\hat z(t')\rangle_0$. These contractions are causal along the sampling direction and vanish at equal times in the It\^o convention, so no such cycle survives.
The normalization $\int\d z_0\,K=1$ of \cref{eq:kernel_normalization} therefore holds order by order in perturbation theory.
This structure shows how nonlinearities in the drift deform free propagation. 
The zeroth-order term is Gaussian transport governed by the linearized dynamics, while higher orders correspond to successive \emph{insertions} of the nonlinear drift $f_{\mathrm{int}}$ at intermediate times.
This insertion structure is precise at the level of probability transport. Splitting the Fokker--Planck generator of \cref{eq:kernel_fpe} according to \cref{eq:drift_split}, the Duhamel formula for a perturbed evolution operator~\cite{ReedSimon1980} gives
\begin{align}
K^{(1)}(z_0,0\mid z_1,1)
=
\int_0^1 \d t \int \d z\;
K^{(0)}(z_0,0\mid z,t)\;
\nabla_z\!\cdot\!\Big(f_{\mathrm{int}}(z,t)\,K^{(0)}(z,t\mid z_1,1)\Big)\eqcomma
\label{eq:kernel_duhamel}
\end{align}
free propagation from the prior to the intermediate time $t$, a single interaction, and free propagation onward, with higher orders iterating this pattern in time-ordered fashion. \Cref{fig:kernel_series} states the same expansion diagrammatically, with $K^{(n)}$ carrying $n$ insertions of $f_{\mathrm{int}}$ whatever that drift happens to be. The interaction vertex is therefore the interacting part of the Fokker--Planck generator, inserted between free kernels. The MSRJD vertex $\imag\,\hat z\cdot f_{\mathrm{int}}$ of \cref{eq:msrjd_interaction} is its path-integral representation, where its diagrammatic counterpart is the vertex field content derived in \cref{sec:diagrammatics}, with exactly one response leg per vertex.

The free MSRJD measure contracts response and latent fields according to the Gaussian propagators defined by $S_0$, generating a systematic series of corrections that can be organized and resummed.
Like perturbative expansions in interacting field theories, this series is in general \emph{asymptotic} rather than convergent, since the number of contractions contributing at order $n$ grows factorially. 
It is reliable as a low-order approximation in the weak-coupling, small-$g^2$ regime~\cite{Zinn-Justin:2002ecy}. 
In this sense, learning a generative model corresponds to tuning the interaction functional $f_{\mathrm{int}}(z,t)$ such that the resulting (interacting) transition kernel in \cref{eq:kernel_perturbative_expansion_full}, when convoluted with the prior as in \cref{eq:scattering_density}, reproduces the observed data distribution.

This scattering viewpoint clarifies the role of determinism and stochasticity in generative modeling.
In the limit $g(t)\to0$, the quadratic noise term
$\tfrac12 g^2(t)\|\hat z\|^2$ in the free MSRJD action vanishes and the response field $\hat z$ becomes a Lagrange multiplier enforcing the deterministic flow, \ie
\begin{align}
\int \mathcal D\hat z\;
\exp\!\left[
-\imag\int_0^1 \d t\;
\hat z(t)\cdot\big(\dot z(t)-f_{\mathrm{rev}}(z(t),t)\big)
\right]
\propto
\delta\!\big[\dot z-f_{\mathrm{rev}}(z,t)\big]\eqperiod
\label{eq:msrjd_g0_delta}
\end{align}
Hence, the path integral collapses onto deterministic trajectories constrained by
\begin{align}
\dot z(t)=f_{\mathrm{rev}}(z(t),t)\eqcomma
\label{eq:msrjd_delta_constraint}
\end{align}
and the transition kernel reduces to its tree-level (saddle-point) approximation, recovering normalizing flows as a leading-order description.
With the response field no longer fluctuating, $S_{\mathrm{int}}$ produces no loop contributions and enters only through the drift $f_{\mathrm{rev}}$ in \cref{eq:msrjd_delta_constraint}.
For finite diffusion strength, the noise term induces fluctuations of the response field and generates higher-order corrections in the expansion of \cref{eq:kernel_perturbative_expansion_full}.

\subsection{Diagrammatic expansion and interpretation}
\label{sec:diagrammatics}

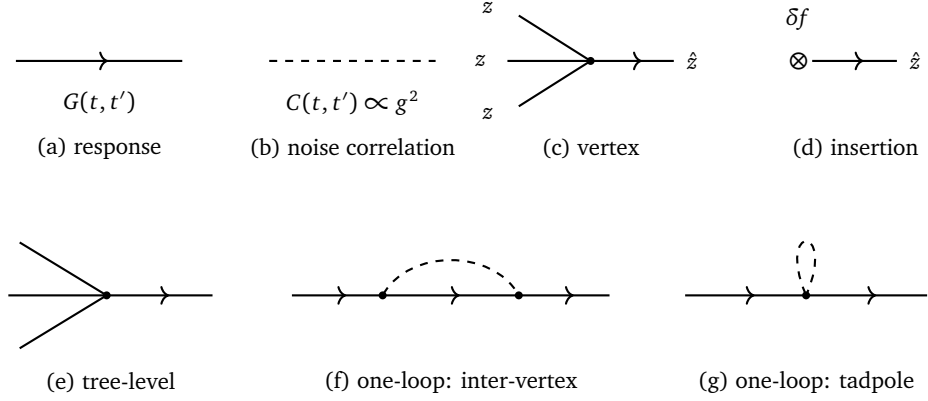
\begin{figure}[t]
\centering
\begin{tikzpicture}[line width=0.9pt, scale=1.0]
\begin{scope}[shift={(0.25,0)}]
  \draw[fermion] (0,0) -- (2.2,0);
  \node at (1.1,-0.55) {\footnotesize $G(t,t')$};
  \node at (1.1,-1.15) {\footnotesize (a) response};
\end{scope}
\begin{scope}[shift={(3.6,0)}]
  \draw[dashed,thick] (0,0) -- (2.2,0);
  \node at (1.1,-0.55) {\footnotesize $C(t,t')\propto g^2$};
  \node at (1.1,-1.15) {\footnotesize (b) noise correlation};
\end{scope}
\begin{scope}[shift={(7.85,0)}]
  \node[vertex] (v) at (0,0) {};
  \draw[fermion] (v) -- (1.1,0);
  \draw[thick] (v) -- (-0.95,0.6);
  \draw[thick] (v) -- (-1.1,0);
  \draw[thick] (v) -- (-0.95,-0.6);
  \node at (1.35,0) {\footnotesize $\hat z$};
  \node at (-1.35,0.7) {\footnotesize $z$};
  \node at (-1.45,0) {\footnotesize $z$};
  \node at (-1.35,-0.7) {\footnotesize $z$};
  \node at (0,-1.15) {\footnotesize (c) vertex};
\end{scope}
\begin{scope}[shift={(10.6,0)}]
  \node at (0,0) {$\otimes$};
  \draw[fermion] (0.18,0) -- (1.3,0);
  \node at (1.55,0) {\footnotesize $\hat z$};
  \node at (0,0.55) {\footnotesize $\delta\!f$};
  \node at (0.75,-1.15) {\footnotesize (d) insertion};
\end{scope}
\begin{scope}[shift={(1.45,-3.1)}]
  \node[vertex] (v2) at (0,0) {};
  \draw[fermion] (v2) -- (1.4,0);
  \draw[thick] (v2) -- (-1.15,0.7);
  \draw[thick] (v2) -- (-1.3,0);
  \draw[thick] (v2) -- (-1.15,-0.7);
  \node at (0.05,-1.15) {\footnotesize (e) tree-level};
\end{scope}
\begin{scope}[shift={(3.9,-3.1)}]
  \draw[fermion] (0,0) -- (1.2,0);
  \node[vertex] (vL) at (1.2,0) {};
  \node[vertex] (vR) at (3.0,0) {};
  \draw[dashed,thick] (vL) to[out=60,in=120] (vR);
  \draw[fermion] (vL) -- (vR);
  \draw[fermion] (vR) -- (4.2,0);
  \node at (2.1,-1.15) {\footnotesize (f) one-loop: inter-vertex};
\end{scope}
\begin{scope}[shift={(9.1,-3.1)}]
  \draw[fermion] (0,0) -- (1.6,0);
  \node[vertex] (w) at (1.6,0) {};
  \draw[dashed,thick] (w) .. controls (1.2,0.95) and (2.0,0.95) .. (w);
  \draw[fermion] (w) -- (3.2,0);
  \node at (1.6,-1.15) {\footnotesize (g) one-loop: tadpole};
\end{scope}
\end{tikzpicture}
\caption{%
Diagrammatic elements of the MSRJD expansion, drawn for representative vertex orders. (a) the response propagator $G(t,t')$ of \cref{eq:response_propagator}. (b) the noise-induced correlation $C(t,t')\propto g^2$ of \cref{eq:zz_propagator}. (c) a vertex from \cref{eq:vertex_family}, here the quartic $\hat z\cdot z^3$ of a cubic drift. (d) the one-point insertion $\imag\hat z\cdot\delta\!f$ of an imperfect learned score. (e) a tree-level contribution of a single vertex. (f) an inter-vertex loop, drawn with the cubic vertices $\hat z\cdot z^2$ of a quadratic drift. (g) a tadpole.
}
\label{fig:probability_flow_diagrams}
\end{figure}

The perturbative expansion of the transition kernel derived in \cref{sec:scattering} organizes into propagators and interaction vertices, in direct parallel with interacting field theories.
Solid lines with a centered arrow denote contracted response propagators, dashed lines denote noise contractions, and plain solid lines denote uncontracted $z$ legs, which attach to external states or to the classical field and turn into an arrowed response line or a dashed noise line only once contracted. The $\hat z$ leg of a vertex or insertion passes its arrow to the response line it emits.
Each insertion of \cref{fig:kernel_series} resolves into vertices once we expand the drift in powers of the state.
Expanding $f_{\mathrm{int}}$ in powers of the state, with couplings written as tensors $\lambda^{(m)}$ of rank $m+1$, the interaction functional in \cref{eq:msrjd_interaction} becomes
\begin{align}
    S_{\mathrm{int}}[z,\hat z]
    =
    \imag \sum_{m\geq2}\int_0^1 \d t\;
    \lambda^{(m)}_{i\,j_1\cdots j_m}(t)\;
    \hat z_i(t)\,z_{j_1}(t)\cdots z_{j_m}(t)\eqcomma
    \label{eq:vertex_family}
\end{align}
with repeated indices summed over the $d$ latent components. The term of order $m$ gives a vertex with one $\hat z$ leg and $m$ legs of $z$. The expansion is used at finite order rather than resummed, so what matters is not its convergence but the size of the couplings left out.
A drift quadratic in $z$ generates the cubic vertex $\hat z\cdot z^2$, a cubic drift the quartic vertex $\hat z\cdot z^3$, and so on. \Cref{fig:probability_flow_diagrams} collects the resulting elements, drawn for representative vertex orders. Which vertices a model carries is fixed by its drift, while the propagators and the counting rules are the same for all of them.

Two features of \cref{eq:vertex_family} are fixed by the construction rather than by the particular drift.
First, every interaction vertex carries at least one response leg $\hat z$.
Integrating $\hat z$ out of \cref{eq:msrjd_action} enforces the equation of motion through $\delta[\dot z - f_{\mathrm{rev}}]$, whose $z$-integral is the normalization in \cref{eq:kernel_normalization} for \emph{any} drift, which a term independent of $\hat z$ would spoil.
With the It\^o Jacobian absorbed into the normalization, the drift nonlinearities generate only single-response vertices $\imag\,\hat z\cdot f_{\mathrm{int}}$, cf.\ the kernel-level statement in  \cref{eq:kernel_duhamel}.
Second, no vertex carries more than two response legs.
The response field $\hat z$ enters quadratically only through the noise term $\tfrac12 g^2\|\hat z\|^2$, so for the state-independent diffusion considered here the sole two-response object is the free noise propagator.
Multiplicative noise $g(z)$ would promote this to a $z$-dependent two-response vertex, and non-Gaussian noise to higher powers of $\hat z$.

The expansion in \cref{eq:kernel_perturbative_expansion_full} is now a sum over diagrams built from response propagators, noise correlations, and the vertices of \cref{eq:vertex_family}.
Tree diagrams involve only response propagators and describe deterministic transport along classical trajectories, while loop diagrams contain at least one $z$--$z$ contraction and carry powers of $g^2$.
In the deterministic limit $g(t)\to0$ the correlations vanish and the expansion truncates at tree level, while at finite diffusion strength the loops contribute the fluctuation corrections that diffusion models sample directly.
Three counting rules fix which diagrams survive at a given order. Every vertex carries exactly one $\hat z$ leg, response fields never contract with each other, and each closed $z$--$z$ contraction costs one power of $g^2$. Together these fix the order of a diagram before any integral is evaluated, while the response propagators fix its causal support. The two examples below apply this to the affine drift, where the expansion terminates, and to a cubic drift, where the first correction appears.

The propagators are the Gaussian averages in \cref{eq:free_average_onesided} over trajectories started from the prior draw with the data endpoint free, and the endpoint delta of \cref{eq:kernel_delta_insertion} enters those averages as one more insertion.
All propagators of the theory are moments and derivatives of the transition kernel. 
The first moment of \cref{eq:scattering_kernel_def} is the conditional mean, and its derivative with respect to the initial state defines the \emph{state-transition Jacobian},
\begin{align}
\mu(t\,|\,z',t') \equiv \int \d z\; z\, K(z,t\mid z',t')
\qquad \mand \qquad
U(t,t') \equiv \frac{\partial \mu(t\,|\,z',t')}{\partial z'}\eqcomma
\label{eq:kernel_first_moment}
\end{align}
the average transport of a state from $t'$ to $t$ and its sensitivity to the starting point. 
In the deterministic limit $U$ is the Jacobian of the flow map $\Phi$, for any drift. 
For an affine drift, $U$ is independent of $z'$ and reduces to the state-transition matrix of the linear dynamics,
\begin{align}
\partial_t U = A(t)\,U
\qquad \mwith \qquad
U(t',t')=\mathbb{I}
\qquad \mand \qquad
U(t,t')=M(t)\,M(t')^{-1}\eqcomma
\label{eq:free_state_transition}
\end{align}
where $M(t)=U(t,0)$ solves the same equation from the origin, cf.\ \cref{eq:jacobian_ode}.
Perturbing the drift instead of the state defines the \emph{response}. Since the generative update subtracts the drift, cf.\ \cref{eq:em_reverse_step}, a shift $b(t')\to b(t')+\delta b(t')$ acts as a state kick $-\delta b(t')\,\d t'$ that is subsequently transported by $U$,
\begin{align}
G(t,t')
\equiv \frac{\delta \langle z(t)\rangle_0}{\delta b(t')}
= -\,\theta(t'-t)\,U(t,t')\eqcomma
\label{eq:response_from_kernel}
\end{align}
causal along the sampling direction. For a nonlinear drift the transported Jacobian becomes state dependent. At equal times, the second cumulant of $K(\,\cdot\,,t\mid z_1,1)$, and across different times, the connected moment of the joint law, which by \cref{eq:chapman_kolmogorov} is a product of kernels,
\begin{align}
\big\langle z(t)\,z(t')^{\!\top} \big\rangle_0
=
\int \d y\,\d y'\;
y\, y'^{\top}\,
K(y,t\mid y',t')\,K(y',t'\mid z_1,1)
\qquad \mfor \qquad t<t'\eqperiod
\label{eq:kernel_joint_moment}
\end{align}
Integrals of the kernel thus give statistics, derivatives give responses, and products give multi-time structure.
The response and the connected correlation can also be expressed as expectation values of \cref{eq:free_average_onesided}:
\begin{enumerate}
\item The first is the \emph{response propagator},
\begin{align}
G(t,t')
\equiv
\big\langle z(t)\,\imag\,\hat z(t')^{\!\top} \big\rangle_0\eqperiod
\label{eq:response_propagator}
\end{align}
Shifting the drift by $b(t)\to b(t)+\delta b(t)$ changes the mean path by
\begin{align}
\delta\langle z(t)\rangle
= \int_0^1\d t'\;G(t,t')\,\delta b(t')\eqperiod
\label{eq:response_relation}
\end{align}
Causality refers to the direction in which the generative dynamics is integrated, from $t=1$ toward $t=0$. A drift perturbation at time $t'$ influences the path only at later stages of sampling, \ie at smaller $t$, so that $G(t,t')=0$ for $t>t'$ and response propagators are inherently ordered along the sampling clock $s=1-t$.

\item The second is the \emph{noise-induced correlation},
\begin{align}
C(t,t')
\equiv
\big\langle z(t)\,z(t')^{\!\top} \big\rangle_0
- \big\langle z(t) \big\rangle_0\big\langle z(t') \big\rangle_0^{\!\top}\eqcomma
\label{eq:zz_propagator}
\end{align}
the connected part of the two-point function. 
The subtraction of the means is essential, as the mean $\langle z(t)\rangle_0$ is the free classical trajectory started at $z_1$ which does not vanish. 
The connected part isolates the noise-induced covariance, which arises entirely from the quadratic noise term $\tfrac12 g^2(t)\|\hat z\|^2$ in the free action, which vanishes in the deterministic limit $g(t)\to0$. 
The response propagator $G(t,t')$ needs no such subtraction because $\langle \hat z\rangle_0=0$, and no independent $\langle \hat z\,\hat z\rangle_0$ propagator exists.
\end{enumerate}
%

\subsubsection*{Example: the affine drift}

The first example switches the interaction off,
\begin{align}
    f_{\mathrm{int}}(z,t) = 0
    \qquad \mand \qquad
    f_{\mathrm{rev}}(z,t) = A(t)\,z + b(t)\eqcomma
    \label{eq:example_affine}
\end{align}
so every $\lambda^{(m)}$ of \cref{eq:vertex_family} vanishes and no vertex exists. The free action is quadratic, so both propagators follow in closed form from the Gaussian averages in \cref{eq:free_average_onesided}. The response propagator inverts the operator $\partial_t - A$ that couples $\hat z$ to $z$ through $\imag\,\hat z\cdot(\dot z - A z)$,
\begin{align}
(\partial_t - A(t))\,G(t,t') = \delta(t-t')\,\mathbb{I}\eqperiod
\label{eq:free_response_greens}
\end{align}
With the causal boundary condition along the sampling direction, the solution is precisely \cref{eq:response_from_kernel}.
The noise term of \cref{eq:msrjd_free} enters the correlation dressed by two response propagators,
\begin{align}
C(t,t')
= \int_0^1 \d u\; G(t,u)\,g^2(u)\,G(t',u)^{\!\top}
= \int_{\max(t,t')}^{1} \d u\; g^2(u)\,U(t,u)\,U(t',u)^{\!\top}\eqcomma
\label{eq:free_propagators_explicit}
\end{align}
using \cref{eq:response_from_kernel} in the last step.
The free kernel then follows from the endpoint insertion in three explicit steps.
First, we write the delta in \cref{eq:kernel_delta_insertion} in its Fourier representation,
\begin{align}
\delta\big(z(0)-z_0\big)
=
\int\frac{\d k}{(2\pi)^d}\;
e^{\imag\,k\cdot(z(0)-z_0)}
=
\int\frac{\d k}{(2\pi)^d}\;
e^{-\imag\,k\cdot z_0}\,
e^{\imag\,k\cdot z(0)}\eqcomma
\label{eq:free_kernel_delta_fourier}
\end{align}
with integration variable $k$.
Inside the free average, $z(0)$ is the fluctuating endpoint of the path, whereas $z_0$ and $k$ are external parameters. By linearity of the average, all path-independent factors pull out,
\begin{align}
K^{(0)}(z_0,0\mid z_1,1)
=
\big\langle\delta\big(z(0)-z_0\big)\big\rangle_0
=
\int\frac{\d k}{(2\pi)^d}\;
e^{-\imag\,k\cdot z_0}\,
\big\langle e^{\imag\,k\cdot z(0)}\big\rangle_0\eqcomma
\label{eq:free_kernel_char_step}
\end{align}
and the remaining average is the characteristic function of the endpoint state.
Second, we evaluate the characteristic function. In the free ensemble the mean path is the classical trajectory. Writing $z(t) = z_\star(t) + \eta(t)$ with $\dot z_\star = A(t)\,z_\star + b(t)$ and $z_\star(1)=z_1$, the affine terms cancel and the free action becomes homogeneous and quadratic in the fluctuations,
\begin{align}
S_0[z_\star+\eta,\hat z]
=
\int_0^1 \d t\;
\left[
\imag\,\hat z\cdot\big(\dot\eta - A(t)\,\eta\big)
+\frac12\,g^2(t)\,\|\hat z\|^2
\right]\eqcomma
\label{eq:free_fluctuation_action}
\end{align}
with $\eta(1)=0$ inherited from the pinned prior end and $\eta(0)$ free.
Three properties of this ensemble carry the evaluation.
The action has no linear term, so the fluctuation has zero mean, $\langle\eta(t)\rangle_0=0$, and the endpoint average is the classical arrival point, $\langle z(0)\rangle_0=z_\star(0)$.
The action is invariant under $(\eta,\hat z)\to(-\eta,-\hat z)$, so all odd moments of $\eta$ vanish.
And since the action is quadratic, the even moments obey Wick's theorem, decomposing into sums over pairwise contractions with the equal-time covariance given by the connected correlator in \cref{eq:zz_propagator}, \ie $\langle\eta(0)\,\eta(0)^{\!\top}\rangle_0 = C(0,0)$.
The endpoint phase then splits into a deterministic factor and a fluctuation average,
\begin{align}
\big\langle e^{\imag\,k\cdot z(0)}\big\rangle_0
=
e^{\imag\,k\cdot z_\star(0)}\,
\big\langle e^{\imag\,k\cdot \eta(0)}\big\rangle_0\eqperiod
\label{eq:free_kernel_phase_split}
\end{align}
Expanding the remaining exponential and averaging term by term, the odd moments drop out, while each even moment of the scalar $k\cdot\eta(0)$ Wick-decomposes into $(2m-1)!!$ pairings, each contributing one factor of $k^{\!\top} C(0,0)\,k$,
\begin{align}
\big\langle e^{\imag\,k\cdot \eta(0)}\big\rangle_0
=
\sum_{m=0}^{\infty}
\frac{(\imag)^{2m}}{(2m)!}\,
\big\langle \big(k\cdot\eta(0)\big)^{2m}\big\rangle_0
=
\sum_{m=0}^{\infty}
\frac{(-1)^m\,(2m-1)!!}{(2m)!}\,
\big(k^{\!\top} C(0,0)\,k\big)^m\eqperiod
\label{eq:free_kernel_wick}
\end{align}
Using $(2m)! = 2^m\, m!\,(2m-1)!!$, the series resums into an exponential,
\begin{align}
\big\langle e^{\imag\,k\cdot \eta(0)}\big\rangle_0
=
\sum_{m=0}^{\infty}
\frac{1}{m!}
\left(-\frac12\,k^{\!\top} C(0,0)\,k\right)^{\!m}
=
\exp\!\Big[-\tfrac12\,k^{\!\top} C(0,0)\,k\Big]\eqcomma
\label{eq:free_kernel_resummed}
\end{align}
and, combining \cref{eq:free_kernel_phase_split,eq:free_kernel_resummed}, the characteristic function is exactly quadratic in $k$,
\begin{align}
\big\langle e^{\imag\,k\cdot z(0)}\big\rangle_0
=
\exp\!\Big[
\imag\,k\cdot z_\star(0)
-\tfrac12\,k^{\!\top} C(0,0)\,k
\Big]\eqperiod
\label{eq:free_kernel_charfun}
\end{align}
Third, inserting \cref{eq:free_kernel_charfun} into \cref{eq:free_kernel_char_step} leaves a Gaussian $k$-integral, which converges since $C(0,0)$ is positive definite for nonvanishing diffusion, and is evaluated by completing the square,
\begin{align}
K^{(0)}(z_0,0\mid z_1,1)
&=
\int\frac{\d k}{(2\pi)^d}\;
\exp\!\Big[
-\imag\,k\cdot\big(z_0-z_\star(0)\big)
-\tfrac12\,k^{\!\top} C(0,0)\,k
\Big]
\notag\\
&=
\frac{
\exp\!\Big[
-\tfrac12\,\big(z_0-z_\star(0)\big)^{\!\top} C(0,0)^{-1}\big(z_0-z_\star(0)\big)
\Big]}
{\sqrt{(2\pi)^d\det C(0,0)}}\notag\\
&=
\mathcal N\big(z_0;\, z_\star(0),\, C(0,0)\big)\eqcomma
\label{eq:free_kernel_gaussian}
\end{align}
the normal density in $z_0$, centered on the endpoint of the classical trajectory $z_\star(0)$, and broadened by the accumulated noise $C(0,0)$. 
In the deterministic limit the covariance vanishes and $K^{(0)}\to\delta\big(z_0-z_\star(0)\big)$, recovering the normalizing-flow transport of \cref{sec:known_models}.
The result is exact. 
With every coupling $\lambda^{(m)}$ equal to zero there is no vertex to insert, only $K^{(0)}$ contributes to \cref{fig:kernel_series}, and the diagrammatic series terminates after its first term.

\subsubsection*{Example: a cubic drift}

The second example keeps the leading odd nonlinearity,
\begin{align}
    f_{\mathrm{int}}(z,t) = \lambda(t)\,z^3\eqcomma
    \label{eq:example_cubic}
\end{align}
so that $\lambda^{(3)}\equiv \lambda(t)$ is the only nonvanishing coupling in \cref{eq:vertex_family}, whose indices we suppress for readability. To first order in $\lambda$, the response propagator is corrected by one insertion of the interaction,
\begin{align}
G(t,t')&\to G(t,t') + G_1(t,t')
\notag\\
\mwith \qquad
G_1(t,t')
&=
-\big\langle z(t)\,\imag\hat z(t')\,S_{\mathrm{int}}[z,\hat z]\big\rangle_{0}
=
\int_0^1\d s\;\lambda(s)\,
\big\langle z(t)\,\imag\hat z(t')\;\imag\hat z(s)\,z(s)^3\big\rangle_{0}\eqcomma
\label{eq:tadpole_first_order}
\end{align}
where no disconnected subtraction is needed since $\langle S_{\mathrm{int}}\rangle_0=0$, as established below \cref{eq:K1_K2_explicit}.
To evaluate the average, we split the latent field into the free mean and its fluctuation, $z=z_\star+\eta$ with $\langle\eta\rangle_0=0$ as in \cref{eq:free_fluctuation_action}, and expand the vertex,
\begin{align}
z(s)^3=z_\star(s)^3+3\,z_\star(s)^2\,\eta(s)+3\,z_\star(s)\,\eta(s)^2+\eta(s)^3\eqperiod
\end{align}
The external background $z_\star(t)$ drops immediately, as in every pairing of the remaining fields one response field is left uncontracted and $\langle\hat z\rangle_0=0$.
Wick's theorem then reduces each term to pairings of $\{\eta(t),\,\imag\hat z(t'),\,\imag\hat z(s),\,\eta(s)^k\}$ built from $G$ and $C$, and most candidates vanish identically. 
Terms with an odd number of fluctuation fields drop by the $(\eta,\hat z)\to(-\eta,-\hat z)$ symmetry of the free action, response fields never pair with each other, $\langle\hat z\,\hat z\rangle_0=0$, and any pairing of $\hat z(s)$ with $\eta(s)$ is an equal-time response, $G(s,s)=0$ in the It\^o convention.
This forces $\hat z(s)$ to pair with the external $\eta(t)$, producing $G(t,s)$, and leaves exactly two surviving patterns. From the term $3z_\star^2\,\eta(s)$, the single fluctuation pairs with $\hat z(t')$, while from $\eta(s)^3$ one of the three legs pairs with $\hat z(t')$ and the remaining two close into the equal-time loop $C(s,s)$. Collecting both,
\begin{align}
G_1(t,t')
=
3\int_0^1 \d s\;
\lambda(s)\,
\Big[\,z_\star(s)^2 + C(s,s)\,\Big]\,
G(t,s)\,G(s,t')\eqcomma
\label{eq:tadpole_example}
\end{align}
with the causal support $t<s<t'$ enforced by the two response propagators.
The two contributions realize the two elementary topologies of \cref{fig:probability_flow_diagrams} and are drawn in \cref{fig:propagator_correction}.

\begin{figure}[t]
\centering
\begin{tikzpicture}[line width=0.9pt, scale=0.9]
  \node at (-1.5,0) {\footnotesize $G_1(t,t')$};
  \node at (-0.55,0) {$=$};
  \begin{scope}[shift={(2.1,0)}]
    \draw[fermion] (-1.5,0) -- (0,0);
    \node[vertex] (v) at (0,0) {};
    \draw[thick] (v) -- (-0.45,0.85);
    \draw[thick] (v) -- (0.45,0.85);
    \node at (-0.62,1.05) {\footnotesize $z_\star$};
    \node at (0.62,1.05) {\footnotesize $z_\star$};
    \draw[fermion] (0,0) -- (1.5,0);
    \node at (-0.95,-0.42) {\footnotesize $G(t,s)$};
    \node at (0.95,-0.42) {\footnotesize $G(s,t')$};
    \node at (2.15,0) {$+$};
  \end{scope}
  \begin{scope}[shift={(6.8,0)}]
    \draw[fermion] (-1.5,0) -- (0,0);
    \node[vertex] (w) at (0,0) {};
    \draw[dashed,thick] (w) .. controls (-0.5,1.15) and (0.5,1.15) .. (w);
    \node at (0,1.28) {\footnotesize $C(s,s)$};
    \draw[fermion] (0,0) -- (1.5,0);
    \node at (-0.95,-0.42) {\footnotesize $G(t,s)$};
    \node at (0.95,-0.42) {\footnotesize $G(s,t')$};
  \end{scope}
\end{tikzpicture}
\caption{%
The two contributions to \cref{eq:tadpole_example}, assembled from the elements of \cref{fig:probability_flow_diagrams}. Both place one vertex at an intermediate time $s$ between two response propagators. On the left the two remaining $z$ legs attach to the classical field and give the tree-level term $\propto z_\star(s)^2$, topology~(e). On the right the same two legs contract with each other into the equal-time correlation $C(s,s)\propto g^2$ and give the tadpole, topology~(g). The choice between these two attachments is what separates tree level from one loop.
}
\label{fig:propagator_correction}
\end{figure}

The background term $\propto z_\star^2$ is of the tree-level type~(e), where two vertex legs attach to the classical field, no noise contraction is involved, and the term survives the deterministic limit $g\to0$. It is a mass insertion, the first order of replacing the affine linearization $A(t)$ by the linearization about the classical trajectory,
\begin{align}
A_{\mathrm{cl}}(t)=A(t)+3\lambda(t)\,z_\star(t)^2\eqcomma
\end{align}
and it is resummed automatically once we expand about the classical path of the full drift, as done in \cref{sec:loop_corrections}.
The tadpole term $\propto C(s,s)\propto g^2$ is the loop element~(g), where two legs of the same vertex contract with each other, the combinatorial factor $3$ counts the pairings, and the contribution vanishes in the deterministic limit.

The two examples illustrate the expansion. The affine drift carries no vertex and the series stops at $K^{(0)}$, while a single cubic coupling already produces both elementary topologies at first order, one surviving the deterministic limit and one carrying the leading power of $g^2$. Both follow from the leg count of the vertex and the two ways its remaining legs attach.
Different classes of generative models correspond to distinct truncations of the same expansion. Normalizing flows retain only tree diagrams, diffusion models evaluate the full series including loops, and conditional flow matching absorbs the fluctuation corrections to the single-time marginals into an effective drift.

\clearpage
\section{Loop corrections to deterministic samplers}
\label{sec:loop_corrections}

Deterministic samplers are fast, which is why they are used. For an exact score the probability-flow ODE even reproduces the time-marginals of the diffusion process exactly.
This is a statement about marginals only. 
The tree-level truncation of the stochastic path integral is the ODE generated by the reverse drift itself, and it misses the trajectory-level and joint statistics of the true reverse process already at exact score. 
Once the score is learned, the marginals deviate as well.

We show that the MSRJD representation organizes the discrepancy as a systematic loop expansion in the diffusion strength $g^2$, derive the leading correction in closed form, requiring two auxiliary equations integrated alongside the deterministic trajectory at no stochastic-sampling cost, and validate it on an exactly solvable model and on nonlinear drifts, including a $24$-dimensional equivariant example.
We further show that an imperfect learned score enters the expansion as a calculable insertion, from which a response-weighted training objective follows as a corollary.

The gap between deterministic and stochastic sampling has recently been approached from a path-integral perspective in Ref.~\cite{hirono2024}, where an interpolating parameter between the probability-flow ODE and the reverse SDE plays the role of Planck's constant and the negative log-likelihood is evaluated in a WKB (semiclassical) expansion in this parameter. 
Our expansion differs in scope and output. It corrects smooth observables of the generated samples rather than the likelihood, its leading correction reduces to two auxiliary ordinary differential equations, and its hybrid reformulation remains well conditioned in the focusing and defocusing regimes near the data manifold.

We work with the MSRJD action from \cref{eq:msrjd_action} and recall the deterministic limit established in \cref{sec:scattering}, where $g(t)\to0$ and integrating out the response field $\hat z$ enforces the constraint
$\dot z = f_{\mathrm{rev}}(z,t)$, and the transition kernel collapses onto the classical trajectory defined by
\begin{align}
    \dot z_{\mathrm{cl}}(t)
    = f_{\mathrm{rev}}\big(z_{\mathrm{cl}}(t),t\big)
    \qquad \mwith \qquad
    z_{\mathrm{cl}}(1) = z_1 \eqperiod
    \label{eq:loop_classical_path}
\end{align}
This drift ODE is the deterministic sampler singled out by the path integral itself, the continuous normalizing flow of \cref{sec:nf_saddle} run with the reverse drift. Initialized at a prior draw $z_1$ and integrated to $t=0$, it returns $z_{\mathrm{cl}}(0)$. 
It is not the probability-flow ODE, which carries the score with coefficient $\tfrac12 g^2$ rather than $g^2$, cf.\ the discussion below \cref{eq:fpe_transport}. 
The offset $\tfrac12 g^2\,\nabla_z\log p$ is itself of order $g^2$ and enters as a calculable insertion of the form derived in
\cref{sec:score_insertions}, so corrections relative to either deterministic sampler follow from the same formulas, with the probability-flow case combining the fluctuation terms below with this tree-level insertion shift.
If the score entering $f_{\mathrm{rev}}$ is \emph{exact}, the probability-flow ODE and the reverse-time SDE share the same time-marginals $p(z,t)$ by construction, and differ only in their trajectory-level and joint statistics. 
In every practical setting, however, the score is replaced by a learned approximation $s_\theta$, the marginals no longer coincide, and the loop expansion becomes an expansion of the \emph{marginal} error as well.
The correction derived below is the leading term in both cases, and the score-mismatch contribution to the marginals is made explicit in \cref{sec:score_insertions}.

\subsection{The fluctuation action}
\label{sec:fluctuation_action}

We expand the latent field about the classical trajectory,
\begin{align}
    z(t) = z_{\mathrm{cl}}(t) + \eta(t)
    \qquad \mwith \qquad \eta(1) = 0\eqcomma
    \label{eq:loop_field_split}
\end{align}
where the boundary condition reflects that the sampler is initialized at a fixed prior draw. This is the interacting counterpart of the free-theory expansion about $z_\star$ in \cref{eq:free_fluctuation_action}, now around the classical trajectory of the full drift. 
Substituting \cref{eq:loop_field_split} into the MSRJD
action and expanding $f_{\mathrm{rev}}$ to first order in $\eta$,
\begin{align}
f_{\mathrm{rev}}(z_{\mathrm{cl}}+\eta,t)
= f_{\mathrm{rev}}(z_{\mathrm{cl}},t)
+ A_{\mathrm{cl}}(t)\,\eta(t)
+ \mathcal{O}(\eta^2)\eqcomma
\qquad
A_{\mathrm{cl}}(t)
\equiv \nabla_z f_{\mathrm{rev}}\big(z_{\mathrm{cl}},t\big)\eqcomma
\label{eq:loop_drift_expansion}
\end{align}
we sort the action by the total power of the fluctuation fields $(\eta,\hat z)$. The field-independent term is a constant, and the term of first order in the fluctuations vanishes because $z_{\mathrm{cl}}$ solves the classical equation of motion in \cref{eq:loop_classical_path}, the saddle-point condition. The leading nontrivial term is therefore the \emph{quadratic} (Gaussian) fluctuation action, which is the free fluctuation action from \cref{eq:free_fluctuation_action} with the affine kernel replaced by the linearization about the classical path, $A\to A_{\mathrm{cl}}$,
\begin{align}
S_2[\eta,\hat z]
= \int_0^1 \d t\;
  \left[
    \imag\,\hat z\cdot\big(\dot\eta - A_{\mathrm{cl}}(t)\,\eta\big)
    + \tfrac12\,g^2(t)\,\|\hat z\|^2
  \right]\eqperiod
\label{eq:loop_quadratic_action}
\end{align}
The interaction functional $S_{\mathrm{int}}[\eta,\hat z]$ collects all terms of cubic and higher order in $\eta$, so by the counting of \cref{sec:diagrammatics} each additional vertex costs a further factor of $g^2$ and \cref{eq:loop_quadratic_action} is the one-loop order of the expansion.
For an affine drift $S_{\mathrm{int}}$ vanishes and this Gaussian order is exact, while for a nonlinear drift one insertion of the leading cubic term still contributes at the same order through the tadpole contraction of \cref{fig:probability_flow_diagrams}\,(g), which we include below.

Because $S_2$ is Gaussian, the fluctuation $\eta$ has zero mean and is fully characterized by its equal-time covariance,
\begin{align}
    C(t) \equiv \big\langle \eta(t)\,\eta(t)^{\!\top}\big\rangle_{S_2}\eqcomma
    \label{eq:loop_cov_def}
\end{align}
where the average runs over fluctuations pinned at the prior end, $\eta(1)=0$, with the endpoint $\eta(0)$ integrated out, as in \cref{eq:free_average_onesided}.
The equal-time covariance is therefore the free correlator with $A\to A_{\mathrm{cl}}$,
\begin{align}
C(t) = \int_t^1 \d u\; g^2(u)\, U_{\mathrm{cl}}(t,u)\,U_{\mathrm{cl}}(t,u)^{\!\top}\eqcomma
\label{eq:loop_cov_integral}
\end{align}
where $U_{\mathrm{cl}}(t,u)$ solves \cref{eq:free_state_transition} with the same replacement.
Differentiating \cref{eq:loop_cov_integral} with respect to $t$ by the Leibniz rule,
\begin{align}
\frac{\d C(t)}{\d t}
&= -\,g^2(t)\,\mathbb{I}
+ \int_t^1 \d u\; g^2(u)\Big[\partial_t U_{\mathrm{cl}}\,U_{\mathrm{cl}}^{\!\top} + U_{\mathrm{cl}}\,\partial_t U_{\mathrm{cl}}^{\!\top}\Big]
\notag\\
&= -\,g^2(t)\,\mathbb{I} + A_{\mathrm{cl}}(t)\,C(t) + C(t)\,A_{\mathrm{cl}}(t)^{\!\top}\eqcomma
\label{eq:loop_cov_derivative}
\end{align}
where $U_{\mathrm{cl}}(t,t)=\mathbb{I}$ fixes the boundary term and $\partial_t U_{\mathrm{cl}}=A_{\mathrm{cl}}U_{\mathrm{cl}}$ the integrand. This is a Lyapunov equation~\cite{Gardiner1985}. Written in components, with summation over repeated latent-space indices implied here and in the following, and with the derivative taken along the sampling direction of decreasing $t$, it reads
\begin{align}
-\frac{\d C_{ij}}{\d t}
= -\big(A_{\mathrm{cl},ik}(t)\,C_{kj} + C_{ik}\,A_{\mathrm{cl},jk}(t)\big)
  + g^2(t)\,\delta_{ij}\eqcomma
\qquad
C_{ij}(1) = 0\eqperiod
\label{eq:loop_lyapunov}
\end{align}
The two terms of \cref{eq:loop_lyapunov} carry signs for different reasons. 
The diffusion term enters with a definite positive sign, because noise variance accumulates along the sampling direction irrespective of the orientation of the time axis. 
The drift term, by contrast, changes sign relative to the forward Lyapunov equation because the derivative is taken with respect to the sampling clock $s = 1-t$. 
With the boundary condition $C(1)=0$, meaning no fluctuations at the initial prior draw, \cref{eq:loop_lyapunov} integrates to a positive semidefinite $C(t)$ for all $t<1$, as a covariance must be.

Let $O$ be an observable of the generated sample, \ie a function of the endpoint state $z(0)$ rather than of the density. Its expectation over the reverse process is the corresponding moment of the endpoint kernel, 
\begin{align}
\langle O\rangle = \int\d z_0\; O(z_0)\,K(z_0,0\mid z_1,1)\eqcomma
\end{align}
which we evaluate perturbatively.
Two effects contribute at order $g^2$. 
First, the Gaussian fluctuations described by \cref{eq:loop_quadratic_action} broaden the endpoint distribution around $z_{\mathrm{cl}}(0)$ with covariance $C(0)$.
Second, for a nonlinear drift the \emph{mean} of the endpoint is shifted at the same order. 
Performing the drift expansion to second order,
\begin{align}
f_{\mathrm{rev},i}(z_{\mathrm{cl}}+\eta,t)
= f_{\mathrm{rev},i}(z_{\mathrm{cl}},t)
+ A_{\mathrm{cl},ij}\,\eta_j
+ \tfrac12\,\partial_j\partial_k f_{\mathrm{rev},i}(z_{\mathrm{cl}},t)\,\eta_j\eta_k
+ \mathcal{O}(\eta^3)\eqcomma
\label{eq:loop_drift_expansion2}
\end{align}
the quadratic term is a cubic vertex in $S_{\mathrm{int}}$ with one $\hat z$ and two $\eta$ legs. 
Read back as a stochastic equation, $S_2$ together with this vertex is the reverse Langevin dynamics of \cref{eq:msrjd_langevin}, linearized about the classical trajectory and carried to second order, so that the fluctuation obeys
\begin{align}
\dot\eta_i = A_{\mathrm{cl},ij}(t)\,\eta_j + \tfrac12\,\partial_j\partial_k f_{\mathrm{rev},i}(z_{\mathrm{cl}},t)\,\eta_j\eta_k + g(t)\,\xi_i\eqcomma
\label{eq:loop_eta_eom}
\end{align}
driven by the same white noise $\xi$ as the reverse process. Averaging over the Gaussian noise, with $\langle\xi_i\rangle=0$ and $\langle\eta_j\eta_k\rangle = C_{jk}$ at leading order (the tadpole contraction of \cref{fig:probability_flow_diagrams}\,(g)), gives a closed equation for the shift $\delta m(t)\equiv\langle\eta(t)\rangle$,
\begin{align}
    -\frac{\d\delta m_i}{\d t}
    = -A_{\mathrm{cl},ij}(t)\,\delta m_j
      - \tfrac12\,
        \partial_j\partial_k f_{\mathrm{rev},i}\big(z_{\mathrm{cl}}(t),t\big)\,
        C_{jk}(t)\eqcomma
    \qquad
    \delta m_i(1) = 0\eqcomma
    \label{eq:loop_mean_shift}
\end{align}
written again with respect to the sampling direction.
The mean shift is sourced by the covariance through the curvature of the drift, and it vanishes identically for affine drifts.
Expanding $O(z_{\mathrm{cl}}(0)+\eta(0))$ to second order and averaging, with $\langle\eta(0)\rangle=\delta m(0)$ and $\langle\eta(0)\,\eta(0)^{\!\top}\rangle=C(0)$ at leading order, gives
\begin{align}
\big\langle O[z(0)]\big\rangle
= O\big(z_{\mathrm{cl}}(0)\big)
+ \partial_i O\big(z_{\mathrm{cl}}(0)\big)\, \delta m_i(0)
+ \tfrac12\,
  \partial_i \partial_j O\big(z_{\mathrm{cl}}(0)\big)\,C_{ij}(0)
+ \mathcal{O}(g^4)\eqcomma
\label{eq:loop_observable_master}
\end{align}
where the first term is the tree-level (deterministic-sampler) value and the remaining two terms constitute the one-loop correction. 
Both are linear in the covariance and hence, by \cref{eq:loop_lyapunov}, linear in $g^2$ at leading
order. 
\Cref{eq:loop_observable_master} is the central formula of this section, the one-loop sampler correction. 
It expresses the leading discrepancy between a deterministic sampler and the true stochastic process through two auxiliary linear equations, \cref{eq:loop_lyapunov,eq:loop_mean_shift}, integrated alongside the classical trajectory.
Propagating the full covariance means carrying $d^2$ components and $\mathcal O(d)$ Jacobian--vector products per step, so large $d$ calls for low-rank or diagonal structure, observable-specific adjoints, or stochastic trace estimation.
Equivalently, it states that the endpoint kernel remains Gaussian at this order, with first and second cumulants $z_{\mathrm{cl}}(0)+\delta m(0)$ and $C(0)$.

The numerical tests below use the second moment $O[z]=\|z\|^2$, the lowest-order observable sensitive to both the mean shift $\delta m(0)$ and the covariance $C(0)$, since a linear observable has vanishing second derivative and reduces \cref{eq:loop_observable_master} to the mean shift alone. For an affine drift that shift vanishes as well, so the free-theory check in \cref{sec:ou_loop_example} rests entirely on $C(0)$. With $\partial_i O = 2z_i$ and $\partial_i\partial_j O = 2\,\delta_{ij}$, the correction becomes
\begin{align}
\big\langle \|z(0)\|^2\big\rangle
= \|z_{\mathrm{cl}}(0)\|^2
+ 2\,z_{\mathrm{cl},i}(0)\,\delta m_i(0)
+ C_{ii}(0)
+ \mathcal{O}(g^4)\eqperiod
\label{eq:loop_second_moment}
\end{align}

\subsection{Imperfect scores as insertions}
\label{sec:score_insertions}

In practice the exact reverse drift is unavailable, since the score entering
$f_{\mathrm{rev}}$ is replaced by a learned approximation $s_\theta(z,t)$, and
the sampler integrates the drift
$f_{\mathrm{rev}}^\theta = f_{\mathrm{rev}} + \delta\!f$ with
\begin{align}
    \delta\!f(z,t)
    = -\,g^2(t)\,\big(s_\theta(z,t)-\nabla_z\log p(z,t)\big)\eqperiod
    \label{eq:score_mismatch_def}
\end{align}
In the MSRJD action the mismatch enters linearly, as the additional term
\begin{align}
    \Delta S[z,\hat z]
    = -\,\imag\int_0^1 \d t\;
      \hat z(t)\cdot \delta\!f(z(t),t)\eqcomma
    \label{eq:score_insertion_action}
\end{align}
a one-point \emph{insertion}, diagrammatically the crossed dot of \cref{fig:probability_flow_diagrams}\,(d) attached to a response line.
Its leading effect is a tree-level shift of the endpoint mean, given by a single contraction of this insertion with the response propagator in \cref{eq:response_propagator},
\begin{align}
    \Delta m(0)
    &= \int_0^1 \d t\;
      G(0,t)\,\delta\!f\big(z_{\mathrm{cl}}(t),t\big)
      + \mathcal{O}\big(\delta\!f^2,\,g^2\,\delta\!f\big)\eqcomma
    \notag\\
    \mwith\qquad
    G(0,t) &= -\,M(0)\,M(t)^{-1}\eqcomma
    \label{eq:score_mean_shift}
\end{align}
with the response propagator evaluated along the classical trajectory in terms of the variational solution $M(t)$ of \cref{eq:jacobian_ode}.
This contribution shifts the learned sampler away from the exact-score marginals, and it is the leading term of the marginal error.
Because \cref{eq:score_mean_shift} is linear in $\delta\!f$, it can also be read in reverse, propagating score-validation errors into uncertainties on generated observables.
The mismatch is also measurable. 
A classifier trained to separate data from model samples at noise level $t$ returns the log density ratio of \cref{eq:gan_optimal_discriminator}, whose gradient is $-\delta\!f/g^2$, and discriminator guidance~\cite{Kim2023dg} trains one at every noise level to estimate this correction and add it back to the learned score during sampling.
The insertion computed here and the correction measured there are the same object, reached analytically in one case and empirically in the other.

\subsubsection*{A response-weighted training objective}

\Cref{eq:score_mean_shift} states which score errors matter, since an error $\delta s(t) \equiv s_\theta-\nabla_z\log p$ committed at time $t$ is transported to the endpoint with amplitude $G(0,t)\,g^2(t)$. 
If score errors at different times are approximately uncorrelated, the induced endpoint mean-squared error is
\begin{align}
    \big\|\Delta m(0)\big\|^2
    \simeq
    \int_0^1 \d t\;
    g^4(t)\,\big\|G(0,t)\,\delta s(t)\big\|^2\eqcomma
\end{align}
so that minimizing the \emph{endpoint} error at leading order, after averaging over error directions, corresponds to the denoising objective in \cref{eq:dsm_weighted} with the weighting
\begin{align}
    \lambda_{\mathrm{resp}}(t)
    \;\propto\;
    g^4(t)\,\big\|G(0,t)\big\|^2\eqcomma
    \label{eq:response_weighting}
\end{align}
the conventional likelihood weighting $g^2(t)$~\cite{Song2021mle} dressed by the squared response propagator, with $\|\cdot\|$ the Frobenius norm.
For the affine reference theory $G(0,t)$ is state independent and available in closed form, $\|G(0,t)\|=\|U(0,t)\|$. For nonlinear drifts it depends on the trajectory, so a usable weighting requires an average over the data flow that we do not evaluate here.
The weighting is also observable aware, and if a specific observable $O$ is targeted, $\|G(0,t)\|^2$ is replaced by $\|\nabla O(z_{\mathrm{cl}}(0))\cdot  G(0,t)\|^2$.
Loss weightings for diffusion models are usually chosen empirically and are known to matter in practice~\cite{Karras2022}. \Cref{eq:response_weighting} provides a first-principles candidate, derived from the propagator structure of the theory.

The same response propagator appears as the adjoint state in reward fine-tuning of diffusion and flow models~\cite{Pan2024adjointdpm,DomingoEnrich2025}, where it transports the gradient of an external objective back through the sampler. 
Here it reweights the score-matching loss itself, so that training effort follows the sensitivity of the generated sample rather than an imposed reward. 
The observable-aware form parallels goal-oriented error estimation, where local residuals carry the weight of an adjoint solution that controls the error in a target functional~\cite{BeckerRannacher2001}.

\subsection{Numerical validation}
\label{sec:ou_loop_example}

We validate the one-loop sampler correction on three models. 
A free theory, where \cref{eq:loop_observable_master} must reproduce the quadratic observable exactly, a low-dimensional nonlinear drift, where it must fail by exactly the next order in $g^2$, and a high-dimensional equivariant drift, which runs the matrix-valued machinery in many dimensions. 
The code reproducing all three studies is publicly available, see the note on code availability at the end of the paper.

\subsubsection*{The Ornstein--Uhlenbeck free theory}

When the reverse drift is affine, $f_{\mathrm{rev}}(z,t)=A(t)\,z+b(t)$, the construction of \cref{sec:free_vs_interacting} identifies the theory as free. 
The action is quadratic, $S_{\mathrm{int}}$ vanishes identically, and the mean shift in \cref{eq:loop_mean_shift} vanishes as well since $\partial_j\partial_k f_{\mathrm{rev},i}=0$. 
The endpoint kernel is then exactly Gaussian, so \cref{eq:loop_observable_master}, which truncates the observable at second order, is exact for observables at most quadratic in $z$. This covers the second moment used throughout, while higher observables pick up Gaussian moments such as $\langle\eta^4\rangle=3C^2$ at order $g^4$.
This provides a
stringent and fully analytic check of the formalism.

For constant $A$ and $b$ the classical trajectory in \cref{eq:loop_classical_path} is a linear relaxation. 
Integrating the linear ODE $\dot z_{\mathrm{cl}} = A\,z_{\mathrm{cl}} + b$ from $z_{\mathrm{cl}}(1)=z_1$ gives
\begin{align}
z_{\mathrm{cl}}(t) = e^{A(t-1)}\,z_1 + \big(e^{A(t-1)}-\mathbb{I}\big)\,A^{-1}b\eqcomma
\label{eq:loop_ou_trajectory}
\end{align}
and evaluating at the data endpoint $t=0$ gives
\begin{align}
z_{\mathrm{cl}}(0) = e^{-A}\,z_1 + \big(\mathbb{I}-e^{-A}\big)\,z_\star
\qquad \mwith \qquad
z_\star = -A^{-1}b\eqcomma
\label{eq:loop_ou_mean}
\end{align}
which relaxes from the prior draw toward the fixed point $z_\star$ of the drift. 
For constant $A$ the state-transition matrix is $U_{\mathrm{cl}}(0,u)=e^{-Au}$, so the covariance integral in \cref{eq:loop_cov_integral} at the endpoint $t=0$ evaluates to
\begin{align}
C(0) = g^2 \int_0^1 \d u\; e^{-A u}\,e^{-A^{\!\top} u}\eqperiod
\label{eq:loop_ou_cov}
\end{align}
A noise kick injected at sampling time $u$ before the endpoint is transported to $t=0$ by the state-transition matrix $e^{-Au}$ and contributes its propagated variance. 
\Cref{eq:loop_ou_cov} is manifestly linear in $g^2$ and positive definite.
The reverse process is in this case an Ornstein--Uhlenbeck process whose endpoint is exactly Gaussian with the mean in \cref{eq:loop_ou_mean} and the covariance in \cref{eq:loop_ou_cov}, so \cref{eq:loop_second_moment} must reproduce the true second moment identically. 
We verify this numerically for the reverse drift $f_{\mathrm{rev}}(z) = A z + b$ in two dimensions, with
\begin{align}
A = \begin{pmatrix} -1.3 & \phantom{-}0.4\\
\phantom{-}0.2 & -0.9 \end{pmatrix}\eqcomma
\qquad
b = \begin{pmatrix} \phantom{-}0.5\\ 
-0.3 \end{pmatrix}\eqcomma
\end{align}
and a fixed prior draw $z_1=(1.0,\,0.5)$.

\begin{figure}[t]
\centering
\includegraphics[width=0.99\linewidth]{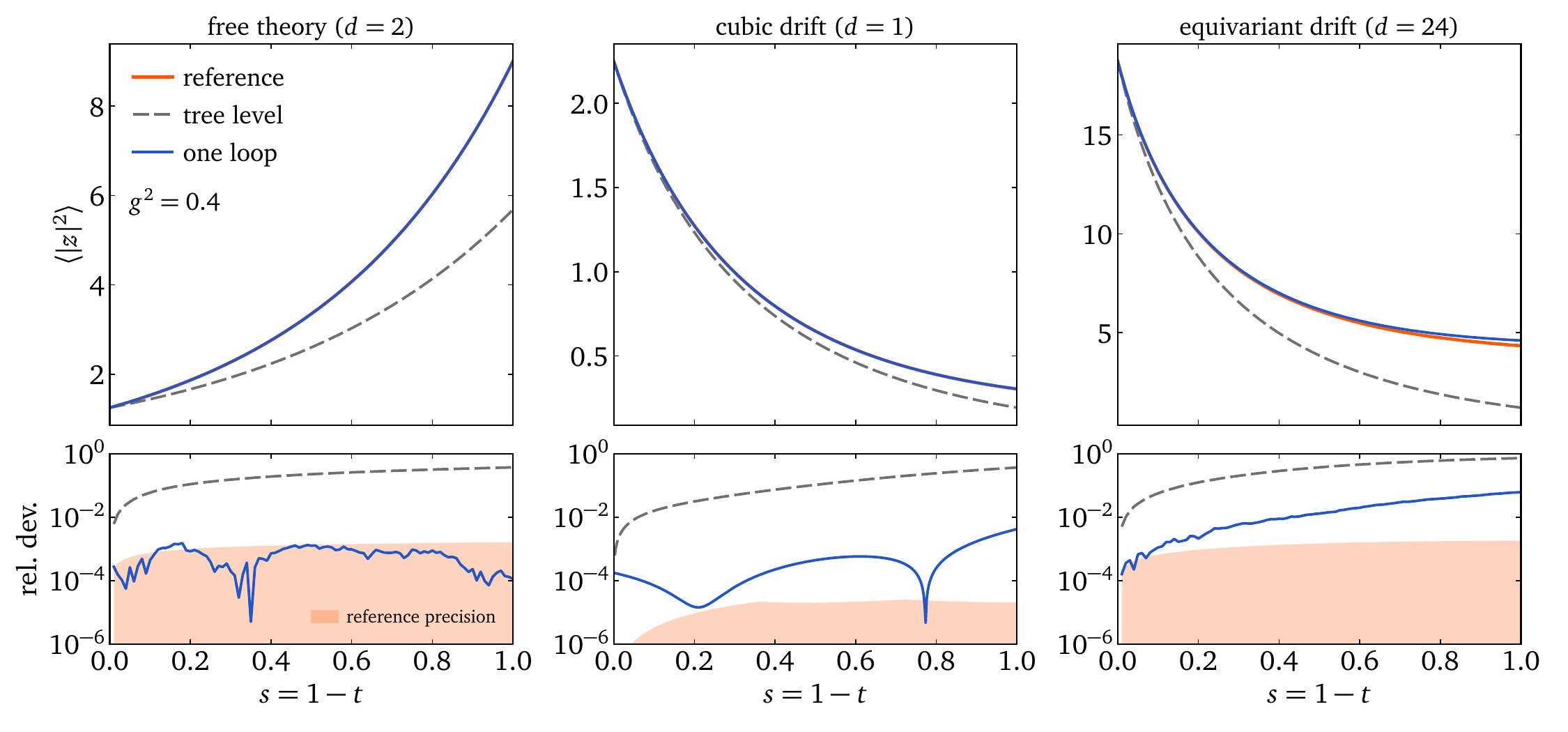}
\caption{%
Transport of the second moment $\langle\|z\|^2\rangle$ along the sampling clock $s=1-t$ for the three validation models, at fixed diffusion strength $g^2=0.4$. Upper panels compare the deterministic sampler at tree level with its one-loop correction against the reference, lower panels give the relative deviation from that reference on a logarithmic scale.
The shaded band gives the precision of the reference itself, statistical for the Euler--Maruyama ensembles of the free and equivariant models and set by grid refinement for the Fokker--Planck solution of the cubic drift.
}
\label{fig:loop_transport}
\end{figure}

Although both the trajectory and the covariance are available in the closed forms just derived, we deliberately run the free theory through the same numerical procedure as the interacting examples, as a check of the solver itself.
The one-loop prediction integrates the classical trajectory in \cref{eq:loop_classical_path} together with the Lyapunov equation~\labelcref{eq:loop_lyapunov} in the sampling clock, using a fourth-order Runge--Kutta scheme with $5\times10^4$ uniform steps. 
Since the drift is affine, the mean shift vanishes and the prediction is $\|z_{\mathrm{cl}}(0)\|^2+\mathrm{Tr}\,C(0)$ with tree-level value $\|z_{\mathrm{cl}}(0)\|^2 = 5.684$.
The reference value is a direct Euler--Maruyama simulation of the reverse SDE, cf.\ \cref{eq:em_reverse_step}, with $2\times10^5$ independent trajectories of $2\times10^3$ steps each, all initialized at the same prior draw $z_1$. 
\Cref{tab:loop_validation} reports, for each diffusion strength, the loop correction $\mathrm{Tr}\,C(0)$, the resulting one-loop prediction, the Monte Carlo reference with its sampling error, and the relative deviation between the two, which is consistent with zero within the quoted errors.

\begin{table}[t]
\setlength{\tabcolsep}{5pt}
\centering
\begin{small}
\begin{tabular}{
  S[table-format=1.1]
  S[table-format=2.3]
  S[table-format=2.3]
  S[table-format=2.3(2)]
  S[table-format=1.2,table-space-text-post={\,\%}]
}
\toprule
{$g^2$} & {$\mathrm{Tr}\,C(0)$} & {one-loop} & {SDE} & {rel.\ error} \\
\midrule
0.2 & 1.660 & 7.344 & 7.345(10) & 0.02\,\% \\
0.4 & 3.320 & 9.004 & 9.003(15) & 0.01\,\% \\
0.6 & 4.980 & 10.664 & 10.659(19) & 0.04\,\% \\
0.8 & 6.641 & 12.324 & 12.315(24) & 0.07\,\% \\
1.0 & 8.301 & 13.984 & 13.971(28) & 0.10\,\% \\
1.4 & 11.621 & 17.305 & 17.281(37) & 0.14\,\% \\
\bottomrule
\end{tabular}
\end{small}
\caption{%
One-loop prediction for the second moment $\langle\|z(0)\|^2\rangle$ versus direct reverse-SDE simulation for the two-dimensional Ornstein--Uhlenbeck free theory, with tree-level value $\|z_{\mathrm{cl}}(0)\|^2 = 5.684$ and $2\times10^5$ Euler--Maruyama trajectories per row. Monte Carlo errors are given in parentheses.}
\label{tab:loop_validation}
\end{table}

Two features of \cref{tab:loop_validation} confirm the analysis. First, the loop correction $\mathrm{Tr}\,C(0)$ is exactly linear in $g^2$, since a linear fit returns a vanishing intercept to machine precision, consistent with \cref{eq:loop_ou_cov}. 
Second, the one-loop prediction matches the stochastic simulation to a relative deviation below $0.15\,\%$ everywhere, within the Monte Carlo sampling noise. 
This is the expected outcome for a free theory, where \cref{eq:loop_observable_master} terminates at one loop, and it validates both the sign structure of the Lyapunov equation~\labelcref{eq:loop_lyapunov} and the observable formula
\labelcref{eq:loop_observable_master}.
The left panels of \cref{fig:loop_transport} show the same statement over the full trajectory at $g^2=0.4$, where the corrected curve tracks the reference within its own sampling precision while the deterministic sampler falls away.

\subsubsection*{An interacting drift}

For a nonlinear drift, \cref{eq:loop_observable_master} is the leading term of a nontrivial series, and both one-loop structures, the covariance and the tadpole mean shift, contribute. We test this on the one-dimensional cubic
reverse drift
\begin{align}
f_{\mathrm{rev}}(z) = a\,z + \lambda\,z^3\eqcomma
\qquad
a = 1.0\eqcomma
\quad
\lambda = 0.3\eqcomma
\quad
z_1 = 1.5\eqcomma
\label{eq:interacting_model}
\end{align}
for which the sampling dynamics $\d z/\d s = -f_{\mathrm{rev}}(z)$ is
globally contracting toward the origin, so that neighbouring trajectories
never cross and the fluctuation problem remains well conditioned over a wide
range of diffusion strengths, cf.\ \cref{sec:loop_caustics}.

The one-loop prediction integrates the classical trajectory, the Lyapunov equation~\labelcref{eq:loop_lyapunov}, and the mean-shift equation~\labelcref{eq:loop_mean_shift} side by side, three scalar equations in one dimension, again with a fourth-order Runge--Kutta scheme on $5\times10^4$ steps. 
We obtain the exact reference independently of any sampling by
solving the Fokker--Planck equation of the reverse process in the sampling clock $s=1-t$,
\begin{align}
\partial_s p(z,s) = \partial_z\big(f_{\mathrm{rev}}(z)\,p(z,s)\big)
+ \tfrac12\,g^2\,\partial_z^2\, p(z,s)\eqcomma
\label{eq:reverse_fpe_numerics}
\end{align}
with an explicit finite-difference scheme on a grid of $3\times10^3$ points covering
$z\in[-3,\,4.5]$, where doubling the grid resolution changes the quoted moments by
less than $4\times10^{-6}$. For consistency with the other examples, the
table also lists a direct Euler--Maruyama simulation with $10^6$
trajectories, which confirms the Fokker--Planck reference within its sampling
errors. Resolving the $\mathcal{O}(g^4)$ residuals themselves, down to
$6\times10^{-5}$, would require some $10^9$ trajectories with
correspondingly refined time steps, which is why the deterministic reference
anchors the relative errors and the scaling fit.
\Cref{tab:loop_interacting} lists, for each diffusion strength, the two
components of the one-loop correction of \cref{eq:loop_second_moment}, the
covariance broadening $\mathrm{Tr}\,C(0)$ and the tadpole mean shift
$2\,z_{\mathrm{cl}}(0)\,\delta m(0)$, together with the resulting one-loop prediction,
the exact Fokker--Planck value, and the relative deviation from it, with and
without the tadpole term.

\begin{table}[t]
\setlength{\tabcolsep}{5pt}
\centering
\begin{small}
\begin{tabular}{
  S[table-format=1.1]
  S[table-format=1.4]
  S[table-format=-1.4]
  S[table-format=1.4]
  S[table-format=1.4]
  S[table-format=1.4(1)]
  S[table-format=1.2,table-space-text-post={\,\%}]
  S[table-format=2.1,table-space-text-post={\,\%}]
}
\toprule
{$g^2$} & {$\mathrm{Tr}\,C(0)$} & {$2\,z_{\mathrm{cl}}\,\delta m(0)$} & {one-loop} & {exact} & {SDE} & {rel.\ error} & {w/o $\delta m$} \\
\midrule
0.1 & 0.0349 & -0.0070 & 0.2202 & 0.2202 & 0.2200(2) & 0.03\,\% & 3.2\,\% \\
0.2 & 0.0698 & -0.0139 & 0.2482 & 0.2480 & 0.2478(2) & 0.10\,\% & 5.7\,\% \\
0.4 & 0.1397 & -0.0278 & 0.3041 & 0.3029 & 0.3026(3) & 0.42\,\% & 9.6\,\% \\
0.6 & 0.2095 & -0.0417 & 0.3601 & 0.3568 & 0.3565(4) & 0.92\,\% & 12.6\,\% \\
0.8 & 0.2793 & -0.0556 & 0.4160 & 0.4095 & 0.4091(5) & 1.58\,\% & 15.2\,\% \\
\bottomrule
\end{tabular}
\end{small}
\caption{%
One-loop sampler correction for the second
moment $\langle z(0)^2\rangle$ versus the exact Fokker--Planck reference for
the interacting model, with tree-level value
$z_{\mathrm{cl}}(0)^2 = 0.1923$. The reference is deterministic and stable
under grid refinement at the $4\times10^{-6}$ level, and the Euler--Maruyama
column confirms it with $10^6$ trajectories, where parentheses denote the
Monte Carlo uncertainty in the last digits. The second and third columns are the two components of the one-loop
correction, covariance broadening and tadpole mean shift, while the last column is
the relative error when the tadpole term is dropped.}
\label{tab:loop_interacting}
\end{table}

A correctly truncated one-loop expansion has to leave a residual of order $g^4$, and this is what \cref{tab:loop_interacting} shows. 
The residual, the relative deviation multiplied by the exact value, gives a log--log slope against $g^2$ of $2.3$ over the full range and $2.1$ over the two smallest values, approaching $2$ from above as $g^2$ decreases, with the excess attributable to $\mathcal{O}(g^6)$ contributions.
Dropping the tadpole degrades the slope to $1.0$. 
The mean shift is therefore not a refinement but part of the one-loop order for nonlinear drifts, and the slope measures the order of the leading term left out.
At the largest diffusion strength shown, the one-loop result reduces a $53\,\%$ tree-level error to $1.6\,\%$.
The center panels of \cref{fig:loop_transport} follow the improvement over the full trajectory at $g^2=0.4$, showing where along the sampling clock the tree-level gap opens.

\subsubsection*{A high-dimensional equivariant drift}

The two tests above establish exactness and scaling in minimal settings, while the formalism itself is matrix-valued and applies in any dimension. 
As a demonstration in many dimensions we take a drift coupling $N=8$ points in $m=3$ dimensions, an interacting theory in $Nm=24$ latent dimensions,
\begin{align}
    f_{\mathrm{rev},i}(z)
    &= a\,z_i + \bar a\,\bar z
    + \lambda_1\,\|z_i\|^2 z_i
    + \lambda_2\,S\,z_i
    + \lambda_3\,(\bar z\!\cdot\! z_i)\,\bar z
    \notag\\
    \mwith \qquad
    \bar z &= \frac1N\sum_j z_j
    \qquad \mand \qquad
    S = \frac1N\sum_j \|z_j\|^2\eqcomma
    \label{eq:equivariant_model}
\end{align}
which is equivariant under permutations of the $N$ points and under simultaneous rotations of all of them.
We set $a = 1.0$, $\bar a = 0.5$, $\lambda_1 = 0.2$, $\lambda_2 = 0.1$ and $\lambda_3 = 0.1$, with a fixed prior draw $z_1\sim\mathcal N(0,\mathbb I)$ and the sampling dynamics again globally contracting.
The one-loop prediction integrates the $24$-dimensional classical trajectory together with the full matrix-valued Lyapunov and mean-shift equations~\labelcref{eq:loop_lyapunov,eq:loop_mean_shift}, with the Jacobian and the Hessian contraction of the drift evaluated exactly. 
The reference is an Euler--Maruyama simulation with $1.3\times10^5$ trajectories and $10^3$ time steps, where doubling the step count shifts the reference by less than $0.01$, negligible against the deviations quoted below. 
At this dimension no grid-based reference is available, and the Monte Carlo precision
suffices for the comparisons below, while the high-precision scaling fit
remains with the one-dimensional model above.

\Cref{tab:loop_equivariant} shows the result. 
The couplings place this model in a strongly fluctuating regime, since already at $g^2=0.1$ the loop correction is comparable to the tree-level value, and at $g^2=0.8$ it exceeds it fivefold, so an uncorrected deterministic sampler misreads the second moment by $41$--$83\,\%$. 
The one-loop prediction reduces this to $1.1\,\%$ at
$g^2=0.1$ and degrades gracefully to $13\,\%$ at the strongest coupling, with the residual growing by a factor of about $3.5$ per doubling of $g^2$, consistent with
the expected $\mathcal{O}(g^4)$ behavior in a regime where higher orders are no
longer negligible, while dropping the tadpole term degrades the deviation by up to a further factor of four. 
The equivariant structure of the drift and the loop machinery of this section thus combine without modification, and the deviation at the strongest coupling marks the onset of the breakdown discussed next.
The right panels of \cref{fig:loop_transport} follow the same model over the full trajectory at $g^2=0.4$, where the one-loop residual stays above the precision of the reference throughout and grows late along the sampling clock, as the $\mathcal{O}(g^4)$ terms accumulate.

\begin{table}[t]
\setlength{\tabcolsep}{5pt}
\centering
\begin{small}
\begin{tabular}{
  S[table-format=1.1]
  S[table-format=1.4]
  S[table-format=-1.4]
  S[table-format=1.4]
  S[table-format=1.4(2)]
  S[table-format=2.2,table-space-text-post={\,\%}]
  S[table-format=2.1,table-space-text-post={\,\%}]
}
\toprule
{$g^2$} & {$\mathrm{Tr}\,C(0)$} & {$2\,z_{\mathrm{cl}}\!\cdot\!\delta m(0)$} & {one-loop} & {SDE} & {rel.\ error} & {w/o $\delta m$} \\
\midrule
0.1 & 0.9186 & -0.0692 & 2.0552 & 2.0336(13) & 1.06\,\% & 4.5\,\% \\
0.2 & 1.8372 & -0.1384 & 2.9046 & 2.8307(20) & 2.61\,\% & 7.5\,\% \\
0.4 & 3.6745 & -0.2767 & 4.6035 & 4.3353(31) & 6.19\,\% & 12.6\,\% \\
0.8 & 7.3490 & -0.5534 & 8.0013 & 7.0519(50) & 13.46\,\% & 21.3\,\% \\
\bottomrule
\end{tabular}
\end{small}
\caption{%
One-loop sampler correction for the second moment $\langle\|z(0)\|^2\rangle$ of the $24$-dimensional equivariant drift, with tree-level value $\|z_{\mathrm{cl}}(0)\|^2 = 1.206$, versus Euler--Maruyama simulation with $1.3\times10^5$ trajectories. 
The parentheses denote the Monte Carlo uncertainty in the last digits. 
The second and third columns are the two components of the one-loop correction, and the last column is the relative error when the tadpole term is dropped.}
\label{tab:loop_equivariant}
\end{table}

\subsection{Caustics and the range of validity}
\label{sec:loop_caustics}

The interacting example above was globally contracting. The first caveat on the range of validity is the size of the loop parameter.
Higher-loop contributions are suppressed by additional powers of $g^2$ weighted by the curvature of the drift, and the one-loop truncation is quantitatively reliable when $g^2\,\|\nabla^2 f_{\mathrm{rev}}\|$ is small along the classical trajectory. 
This is the regime of low-to-moderate diffusion, which includes near-deterministic samplers.

The second is more subtle and is specific to the generative setting. 
The linearized operator $A_{\mathrm{cl}}(t)$ in \cref{eq:loop_drift_expansion} inherits, through the score term in \cref{eq:rev_sde}, the curvature of the log-density, which grows without bound as the marginals sharpen toward the data manifold. 
Two degenerate regimes result, distinguished by the sign of this curvature along a given direction. 
Along directions that contract onto a sharply peaked mode, the sampling flow \emph{focuses}, \ie neighbouring trajectories are driven together and the map from prior to data degenerates in the limit.
At any finite time the flow map of a smooth drift remains invertible, and the degeneracy is only approached in the limit.
This focusing is the probability-flow analogue of a
\emph{caustic} in optics, where a caustic is the envelope on which a family of light rays is driven together and the ray density diverges, familiar as the bright lines on the bottom of a swimming pool. 
Here the rays are the classical trajectories of the probability flow.
The corresponding eigenvalues of $C(t)$ are squeezed toward the small equilibrium value set by the balance of contraction and noise, the precision $C(t)^{-1}$ becomes large, and the propagation of \cref{eq:loop_lyapunov} becomes numerically stiff.
Along directions in which neighbouring trajectories instead separate toward different modes, near the boundaries between basins of attraction, the flow \emph{defocuses}, and the corresponding eigenvalues of $C(t)$ grow exponentially. 
In each regime one of the two matrices $C$ or $C^{-1}$
degenerates while the other remains well conditioned. 
The free Ornstein--Uhlenbeck theory of \cref{sec:ou_loop_example} has a constant, curvature-free linearization, so neither degeneracy arises, which is why
the one-loop result is there exact and well behaved.

\subsubsection*{Regular reformulation via a Riccati equation}

Both degeneracies are artifacts of committing to a single variable, not of the underlying physics. 
They are removed by propagating, alongside the covariance, its inverse, the precision matrix
\begin{align}
    P(t) \equiv C(t)^{-1}\eqperiod
    \label{eq:loop_precision_def}
\end{align}
Writing $s=1-t$ for the sampling clock, the Lyapunov equation~\labelcref{eq:loop_lyapunov} reads
\begin{align}
\frac{\d C_{ij}}{\d s}
= -\big(A_{\mathrm{cl},ik}\,C_{kj} + C_{ik}\,A_{\mathrm{cl},jk}\big)
  + g^2\,\delta_{ij}\eqcomma
\label{eq:loop_lyapunov_s}
\end{align}
and differentiating $P = C^{-1}$ through the identity
$\d P/\d s = -P\,(\d C/\d s)\,P$ gives the matrix Riccati equation
\begin{align}
\frac{\d P_{ij}}{\d s}
= P_{ik}\,A_{\mathrm{cl},kj}(s) + A_{\mathrm{cl},ki}(s)\,P_{kj}
  - g^2(s)\,P_{ik}\,P_{kj}\eqcomma
\label{eq:loop_riccati}
\end{align}
quadratic in the unknown matrix through its last term.
The two equations are complementary. 
A rapidly growing eigenvalue of $C$, encountered along defocusing directions, corresponds to an eigenvalue of $P$ approaching zero. 
The right-hand side of \cref{eq:loop_riccati} is polynomial in $P$ with bounded coefficients, so the precision formulation stays regular in this limit even where propagation in $C$ becomes poorly conditioned.
Conversely, in the focusing regime it is $P$ that becomes large, while the Lyapunov equation for $C$ remains regular as the corresponding eigenvalues of $C$ become small.
In particular, the boundary condition $C(1)=0$ itself corresponds to $P\to\infty$, so the propagation must in any case start from the Lyapunov form. 
The stable procedure is therefore hybrid. 
We monitor the eigenvalue spectrum of $C(t)$ and propagate whichever of $C$ or $P$ is currently well conditioned, switching back when the spectrum recovers. 
Because the two descriptions are exact inverses, the switch introduces no approximation.
By Jacobi's formula,
\begin{align}
\frac{\d}{\d s}\log\det C = P_{ij}\,\frac{\d C_{ji}}{\d s}\eqcomma
\label{eq:loop_jacobi}
\end{align}
and inserting the Lyapunov equation~\labelcref{eq:loop_lyapunov_s} gives
\begin{align}
    \frac{\d}{\d s}\log\det C(s)
    = -2\,A_{\mathrm{cl},kk}(s)
      + g^2(s)\,P_{kk}(s)\eqcomma
    \label{eq:loop_det_evolution}
\end{align}
which is finite in the defocusing regime, where $P\to0$ even as eigenvalues of $C$ diverge. In the focusing regime its decrease reflects the genuine sharpening of the endpoint distribution rather than a breakdown of the method. The hybrid propagation of \cref{eq:loop_lyapunov,eq:loop_riccati}, together with \cref{eq:loop_det_evolution}, thus remains well conditioned whenever one regime dominates the spectrum, as in the examples above. When focusing and defocusing coexist along different directions, neither matrix is globally well conditioned, and a factored square-root propagation of $C$ extends the scheme to the mixed case.

At order $g^2$ the correction has three pieces, the Gaussian broadening, the tadpole mean shift, and the score insertion, and each reduces to an ordinary differential equation integrated alongside the classical trajectory. 
The correction needs only Jacobian and Hessian products of the drift, at $\mathcal O(d)$ products per step for the full covariance.
The low-rank and diagonal approximations of \cref{sec:fluctuation_action} carry it to large $d$.

\clearpage
\section{Symmetry-constrained flows from EFT power counting}
\label{sec:eft_power_counting}

\Cref{sec:loop_corrections} organized fluctuations around a Gaussian reference. The drift itself was left arbitrary.
In an effective field theory the interaction terms are not arbitrary. They are the most general local operators consistent with the symmetries of the problem, ordered by a power-counting scheme that ranks their importance.
We apply the same logic to the reverse drift $f_{\mathrm{rev}}(z,t)$, which turns the choice of architecture from an empirical matter into an operator enumeration.

\subsection{The drift as an effective expansion}
\label{sec:eft_expansion}

The interaction functional $S_{\mathrm{int}}[z,\hat z]$ of
\cref{eq:msrjd_interaction} is linear in the drift by construction, since
$f_{\mathrm{int}}$ appears contracted with a single response field. Any structure
imposed on $f_{\mathrm{int}}$ therefore passes directly into the vertices of the
diagrammatic expansion. We fix this structure by the same two ingredients
that fix an effective Lagrangian, a symmetry and a power-counting scheme.

Let $G$ be a group acting on the latent space through a representation
$\rho(\group)$, $\group\in G$. We require the generative model to be
$G$-equivariant, so that if the data distribution is invariant,
\begin{align}
    p_{\mathrm{data}}(\rho(\group)\,z) = p_{\mathrm{data}}(z)
    \qquad \forall\,\group\in G\eqcomma
    \label{eq:eft_data_invariance}
\end{align}
the generated distribution should be invariant as well. At the level of the
dynamics this is guaranteed if the reverse drift is $G$-equivariant,
\begin{align}
    f_{\mathrm{rev}}(\rho(\group)\,z, t)
    = \rho(\group)\,f_{\mathrm{rev}}(z,t)
    \qquad \forall\,\group\in G\eqcomma
    \label{eq:eft_equivariance}
\end{align}
and the prior and diffusion are $G$-invariant. 
\Cref{eq:eft_equivariance} restricts $f_{\mathrm{rev}}$ to the space of equivariant vector fields on the latent space, and the free/interacting split in \cref{eq:drift_split} respects this, since the affine part and the
nonlinear remainder are separately constrained.

A symmetry alone leaves an infinite-dimensional space of admissible drifts.
The second ingredient is a power-counting scheme that orders this space.
We assign to each candidate operator a \emph{degree} under the rescaling
\begin{align}
    z \;\to\; \lambda\,z
    \qquad \mand \qquad
    t \;\to\; t\eqcomma
    \label{eq:eft_scaling}
\end{align}
which is the natural dilation of latent space, and we count powers of the diffusion strength $g$.
An operator $\mathcal{O}(z,t)$ contributing to the drift that scales as $\mathcal{O}(\lambda z,t)=\lambda^{n}\mathcal{O}(z,t)$ is said to have degree $n$. 
The affine drift contains the degree-one operator $z$ and, when the symmetry admits it, a constant, while the leading interactions are the lowest-degree equivariant operators not already present in the free theory.

This ordering by degree is equivalently a dimensional analysis. Requiring the MSRJD action in \cref{eq:msrjd_action} to be invariant under the dilation \labelcref{eq:eft_scaling}, with $t$ held fixed, assigns the weights 
\begin{align}
  [z]=1\eqcomma \qquad [\hat z]=-1\eqcomma \qquad \mand \qquad [g]=1\eqcomma
\end{align}
so that the response propagator and covariance satisfy
\begin{align}
    [G]=0 \qquad \mand \qquad [C]=2\eqcomma
\end{align}
as required. 
These weights are fixed by the free theory rather than chosen, which is what makes the counting more than a convention.
A degree-$n$ drift operator then enters with a coupling of weight $1-n$, so the degree-one coefficients are dimensionless while the leading degree-three couplings carry weight $-2$.
Weights alone do not order anything. They become a hierarchy once measured against a scale, so we introduce a latent scale $\Lambda(t)$ and write
\begin{align}
    f_{\mathrm{rev}}(z,t)
    = \sum_{n,a} c^{(n)}_a(t)\,\Lambda^{1-n}(t)\,\mathcal{O}^{(n)}_a(z)
    \label{eq:eft_scaled_expansion}
\end{align}
where $n$ runs over operator degrees, $a$ labels the independent $G$-equivariant operators at each degree, and the $c^{(n)}_a$ are dimensionless. Two lengths control the expansion,
\begin{align}
    \ell(t) \equiv \sqrt{\big\langle\|z(t)\|^2\big\rangle}
    \qquad \mand \qquad
    \Lambda(t) \equiv
    \frac{\big\|\nabla f_{\mathrm{rev}}\big\|}{\big\|\nabla^2 f_{\mathrm{rev}}\big\|}
    \bigg|_{z_{\mathrm{cl}}(t)}\eqcomma
    \label{eq:eft_scales}
\end{align}
where $\ell$ is the typical size of a latent configuration and $\Lambda$ is the distance over which the drift departs from linearity. The second is a curvature scale in the ordinary sense, the length on which the second derivative of $f_{\mathrm{rev}}$ becomes comparable to the first, and it is the same object that controls the loop parameter in \cref{sec:loop_caustics}.
Degree-$n$ operators are suppressed by $(\ell/\Lambda)^{n-1}$ when the dimensionless coefficients are of order unity, so the expansion is controlled whenever $\ell(t)\ll\Lambda(t)$, meaning the configuration stays within the region where the drift is close to linear. Without this separation the degree remains a well-defined ordering but carries no error estimate.

Whether operator degree and loop order are independent axes depends on the regime. The latent size of \cref{eq:eft_scales} splits as $\ell^2 = \|z_{\mathrm{cl}}\|^2 + \tr C$ with $C\propto g^2$, so when the classical trajectory dominates, $\ell/\Lambda$ carries no dependence on $g$ and the two expansions factorize. A higher-degree operator then enters at tree level suppressed by $\ell/\Lambda$ alone, while each loop costs $g^2$ regardless of the degree of its vertices.
When fluctuations dominate instead, $\ell\propto g$ and each factor of $(\ell/\Lambda)^2$ is itself a factor of $g^2/\Lambda^2$. Degree and loop counting then collapse onto a single parameter, as in the Weinberg counting of chiral effective field theory where every loop costs $(p/\Lambda_\chi)^2$~\cite{Weinberg1979,Weinberg1990}. The near-deterministic samplers of \cref{sec:loop_corrections} sit in the first regime, where the double expansion is genuine.
For models built from permutation-invariant pooling a third axis appears, the inverse point number $1/N$, which enters through the intensive normalization of the pooled operators and the multiplicity of the relative modes. 
We do not develop this large-$N$ counting here, but it is the natural parameter organizing the expansion at large point number, and it is what renders the free-theory loop blocks of \cref{sec:eft_permutation} independent of $N$.

The resulting structure is that of a \emph{worldline} effective field theory, in which the latent state traces out a trajectory -- a worldline -- in latent space, parametrized by the auxiliary time $t$, and the drift operators are local operators living on this worldline, organized by symmetry and power counting. 
The construction closely parallels the worldline EFTs used for extended objects in gravity, where a compact body
is reduced to a point particle dressed with a tower of symmetry-allowed operators on its worldline, with coefficients fixed by
matching~\cite{Goldberger:2004jt,Porto:2016pyg}. Here the latent configuration plays the role of the point particle, and training performs the matching.

Given a symmetry group $G$ and a target accuracy, one enumerates the $G$-equivariant operators up to a chosen maximum degree, retains them with free coefficients, and discards the rest. The retained coefficients are the learnable parameters of the model, and for $\ell\ll\Lambda$ the truncation error is set by the degree of the first neglected operator.

\subsection{Permutation-equivariant drift}
\label{sec:eft_permutation}

Consider latent configurations consisting of $N$ points in $\mathbb{R}^{m}$,
\begin{align}
    z = (z_1,\dots,z_N)
    \qquad \mwith \qquad z_i\in\mathbb{R}^{m}\eqcomma
\end{align}
and take the symmetry group to be $G = S_N\times O(m)$.
Throughout this section, subscripts on $z$ label the points of the configuration, not times, as the trajectory endpoints $z(0)=z_0$, $z(1)=z_1$ do not appear here.
The symmetric group $S_N$ permutes the points, while $O(m)$ rotates and reflects each point simultaneously,
\begin{align}
    \rho(\sigma)\,z
    = \big(z_{\sigma^{-1}(1)},\dots,z_{\sigma^{-1}(N)}\big)
    \qquad \mand \qquad
    z_i \;\to\; R\,z_i
    \qquad \mwith \qquad
    R \in O(m)\eqperiod
    \label{eq:eft_group_action}
\end{align}
This is the relevant symmetry for generative models of sets of isotropic data, from point clouds and collections of particles to the constituents of a jet in the high-energy-physics setting
that motivates this work, where the physics is invariant under reordering of the constituents and, for suitable coordinates, under rotations. 
Equivariance requires the drift on point $i$ to be a symmetric function of the other points that transforms as a vector under $O(m)$.
The rotation factor has an immediate structural consequence. Since $-\mathbb{I}\in O(m)$, the drift must be odd,
\begin{align}
    f_{\mathrm{rev}}(-z,t) = -f_{\mathrm{rev}}(z,t)\eqcomma
    \label{eq:eft_parity}
\end{align}
so equivariant operators exist only at odd polynomial degree and a constant term is excluded. 
Consistently, the prior must be $G$-invariant, as is the standard normal. 
The absence of even-degree operators is thus not an accident of the enumeration below but a prediction of the symmetry.
Equivariant architectures for such data are well established, from permutation-invariant function representations like deep sets~\cite{Zaheer2017}, to equivariant flows and diffusion models~\cite{Kohler2020,Satorras2021,Hoogeboom2022} and their applications to particle clouds in collider physics~\cite{Buhmann:2023pmh,Leigh:2023toe}. 
What the field-theoretic construction adds is not equivariance itself but its organization.

The equivariant operators are built from the per-point vector $z_i$ together with permutation-invariant ``pooling'' combinations of the configuration. Completeness at each degree follows from the first fundamental theorem for $O(m)$~\cite{Weyl1946}, by which every $O(m)$-invariant of a set of vectors is a function of their inner products, so that every equivariant vector field is such an invariant multiplying $z_i$ or a pooled vector. 
Up to degree three, the relevant building blocks are
\begin{align}
\bar z \equiv \frac1N\sum_j z_j\eqcomma
\qquad
S \equiv \frac1N\sum_j \|z_j\|^2\eqcomma
\qquad
T \equiv \frac1N\sum_j z_j z_j^{\!\top}\eqcomma
\qquad
w \equiv \frac1N\sum_j \|z_j\|^2\,z_j\eqcomma
\label{eq:eft_building_blocks}
\end{align}
of degrees one, two, two, and three respectively. Organized by degree, the equivariant vector fields acting on point $z_i$ are
\begin{align}
\text{degree }1:\quad & z_i\eqcomma \qquad \bar z\eqcomma
    \nonumber\\[0.4ex]
\text{degree }3:\quad & \|z_i\|^2 z_i\eqcomma \qquad
    S\,z_i\eqcomma \qquad
    (\bar z\!\cdot\! z_i)\,\bar z\eqcomma \qquad \dots
\label{eq:eft_operator_list}
\end{align}
where the complete degree-three basis contains eleven operators, obtained from all $O(m)$-contractions of three vectors drawn from $\{z_i,\bar z\}$ and the pooled moments~\labelcref{eq:eft_building_blocks}, the six structures $\|z_i\|^2 z_i$, $(\bar z\!\cdot\!z_i)\,z_i$, $\|\bar z\|^2 z_i$, $\|z_i\|^2\,\bar z$, $(\bar z\!\cdot\!z_i)\,\bar z$, $\|\bar z\|^2\,\bar z$, the four moment-dressed structures $S\,z_i$, $S\,\bar z$, $T\,z_i$, $T\,\bar z$, and the pooled vector $w$. For small $N$ or $m$ some of these structures become linearly dependent, \eg $T$ coincides with $S$ for $m=1$.
The degree-one operators reproduce, after summation with free coefficients, the most general linear equivariant drift, a self term $\propto z_i$ and a mean-field term $\propto\bar z$. 
This is the free theory of \cref{sec:free_vs_interacting}, now identified as the degree-one truncation of the equivariant expansion. The degree-three operators are the leading
interactions, and each is cubic in $z$ and hence generates a four-point vertex of the MSRJD expansion, \ie one response leg and three latent legs, see \cref{fig:probability_flow_diagrams}\,(c).

Retaining the complete basis up to degree three with time-dependent coefficients defines the leading equivariant reverse drift,
\begin{align}
f_{\mathrm{rev},i}(z,t)
= \underbrace{
    a(t)\,z_i + \bar a(t)\,\bar z
  }_{\text{free, degree }1}
+ \underbrace{
    \lambda_1(t)\,\|z_i\|^2 z_i
    + \lambda_2(t)\,S\,z_i
    + \lambda_3(t)\,(\bar z\!\cdot\! z_i)\,\bar z
    + \ldots
  }_{\text{leading interactions, degree }3}\eqcomma
\label{eq:eft_leading_drift}
\end{align}
where all scalar coefficients are functions of $t$ only. Truncated at degree three and with constant coefficients, this is the drift used for the $24$-dimensional test in \cref{sec:ou_loop_example}.
The power counting makes the role of each term transparent. 
The free coefficients $a,\bar a$ define the Gaussian reference flow and are fixed, as in \cref{sec:loop_corrections}, by the linearized dynamics.
The couplings $\lambda_k$ multiply the leading interaction vertices, distinguished by how the four legs are routed through the point index. They are the independent components that the symmetry leaves in the rank-four tensor $\lambda^{(3)}$ of \cref{eq:vertex_family}, reduced from $\mathcal{O}\big((Nm)^4\big)$ free entries to eleven numbers.
Operators of degree five and higher are the first neglected terms, and their omission is controlled when $\ell\ll\Lambda$ and the dimensionless coefficients stay of order unity.

\subsubsection*{Consequences of the power counting}

The construction changes what a network has to learn, as we can see for score matching and conditional flow matching. 
Score matching and conditional flow matching regress different objects, but the forward process fixes $f(z,t)$ and $g(t)$, so both targets follow from $f_{\mathrm{rev}}$ alone,
\begin{align}
    s(z,t) &\equiv \nabla_z\log p(z,t)
    = \frac{1}{g^2(t)}\Big[\,f(z,t) - f_{\mathrm{rev}}(z,t)\,\Big]\eqcomma
    \nonumber\\[0.4ex]
    v(z,t) &= \tfrac12\Big[\,f(z,t) + f_{\mathrm{rev}}(z,t)\,\Big]\eqcomma
    \label{eq:eft_score_velocity}
\end{align}
the score of \cref{eq:rev_sde} and the probability-flow velocity of \cref{eq:probability_flow_ode}.
Replacing $f_{\mathrm{rev}}$ by the truncated expansion in \cref{eq:eft_leading_drift}, written $f^{\mathrm{EFT}}_\theta$ with coefficient functions $a(t)$, $\bar a(t)$, and $\lambda_k(t)$ parametrized by a network with parameters $\theta$, gives
\begin{align}
    s_\theta(z,t) &= \frac{1}{g^2(t)}\Big[\,f(z,t) - f^{\mathrm{EFT}}_\theta(z,t)\,\Big]\eqcomma
    \nonumber\\[0.4ex]
    v_\theta(z,t) &= \tfrac12\Big[\,f(z,t) + f^{\mathrm{EFT}}_\theta(z,t)\,\Big]\eqperiod
    \label{eq:eft_ansatz}
\end{align}
Standard score matching fits a generic $s_\theta(z,t)$ and enforces the symmetry through the architecture.
Here the functional form is fixed, so the network maps a single time argument to thirteen coefficients rather than an $(N\times m)$-dimensional state to an $(N\times m)$-dimensional output, and the losses in \cref{eq:dsm_weighted,eq:cfm_loss} apply unchanged.
Equivariance holds identically because $f^{\mathrm{EFT}}_\theta$ is assembled from equivariant operators, once the prior and the conditional interpolation share the symmetry of the data.
The score ansatz divides an $\mathcal{O}(g^2)$ difference by $g^2$, so errors in $f^{\mathrm{EFT}}_\theta$ are amplified at small $g$, while the velocity ansatz is free of this.

First, the expansion is \emph{equivariant order by order}. 
Because the vertices in \cref{eq:eft_leading_drift} and the propagators of the equivariant free theory are separately equivariant, every diagram built from them is equivariant, and the symmetry is preserved order by order in $g^2$ as an exact property of the truncation rather than an approximate property of a trained model. The closure holds for the symmetry and not for the degree, since higher orders can generate symmetry-allowed structures of higher degree, which the expansion then orders in the same way.

Second, the scheme \emph{predicts a hierarchy}. The relative weight of the degree-three interactions against the linear flow measures $(\ell/\Lambda)^2$ directly,
\begin{align}
\epsilon(t) \equiv
\frac{\max_k |\lambda_k(t)|\,\ell^2(t)}{|a(t)|}\eqcomma
\label{eq:eft_expansion_parameter}
\end{align}
and, under the same order-unity assumption for the dimensionless coefficients, degree-$(2k{+}1)$ operators contribute at relative order $\epsilon^{k}$.
The parameter is invariant under rescalings of the latent space, since the degree-three couplings carry two inverse powers of $\Lambda$, and it is a measurable property of the fitted model rather than a schedule choice.
This makes the hierarchy falsifiable. Fitted degree-five couplings should be suppressed relative to fitted degree-three couplings by a further power of $\epsilon$, and a measured $\epsilon$ of order unity would show that the truncation is not controlled for that data.

Third, the symmetry \emph{reduces the cost of the loop expansion} of \cref{sec:loop_corrections}. 
For the free part of \cref{eq:eft_leading_drift},
the $S_N$ isotypic decomposition~\cite{FultonHarris1991} -- the decomposition of the linearized dynamics into its irreducible symmetry sectors -- splits the $(N\times m)$-dimensional linearized dynamics into the mean mode $\bar z$, with drift eigenvalue $a+\bar a$, and $N{-}1$ identical relative modes with eigenvalue $a$. 
The Lyapunov equation~\labelcref{eq:loop_lyapunov} then factorizes into two $m\times m$ blocks, independent of $N$. 
Around a generic, non-symmetric classical trajectory the interacting linearization respects this block structure only approximately, but it remains the natural preconditioner for the one-loop computation at large $N$.

The combination of the equivariance condition in \cref{eq:eft_equivariance} and the power counting in \cref{eq:eft_scaling} selects a finite, ordered set of operators from the infinite-dimensional space of admissible drifts, replacing architecture search with operator enumeration.
The same construction applies to any symmetry for which the equivariant operators can be enumerated -- Lorentz symmetry for the four-momenta of high-energy physics, or $S_N$ alone when no rotational structure is present, in which case additional even-degree operators such as componentwise products appear -- with the group changing only the operator list \labelcref{eq:eft_operator_list}, not the logic of the expansion. 
The measured $\epsilon$ of a fitted model decides whether the truncation is controlled for a given dataset, which makes \cref{eq:eft_leading_drift} a hypothesis with a built-in test rather than an architecture choice.

\clearpage
\section{Outlook}
\label{sec:outlook}

Path integrals organize generative models into a single master action, and that organization turns sampler errors, score imperfections, and architecture design into perturbative computations.
The resulting picture interprets generative modeling as a worldline effective field theory on probability space, in the sense of \cref{sec:eft_power_counting}. 
Latent variables evolve along an auxiliary time direction, an Onsager--Machlup action weighs their trajectories, and the common generative model classes occupy the familiar regimes of such a theory. 
Normalizing flows are its classical limit, in which the path integral collapses onto a single trajectory and exact likelihoods follow from a change of variables.
Finite diffusion turns on fluctuations around this classical path, and affine drifts single out the free theory.
Nonlinear diffusion models are then interacting worldline theories.
Their exact likelihoods are generally intractable, common likelihood approximations truncate the underlying functional integral, and score-based training implicitly reconstructs the full interacting drift.

Conditional flow matching fits the same picture. It is not a distinct probabilistic model but an evaluation principle that replaces stochastic sampling by a deterministic flow, retains the finite-noise marginals, and absorbs the unsampled fluctuations into an effective transport field.
Which parts of the interacting expansion it retains and which it averages over, compared with score matching and Schr\"odinger bridges, ties empirical performance to the diagrammatic structure. The scattering viewpoint of \cref{sec:scattering} treats conditioning, partial observations, and constraints as operator insertions, which places conditional generation in the same formalism.

The loop expansion of \cref{sec:loop_corrections} carries over to trained diffusion models without modification, since the hybrid Lyapunov--Riccati formulation of \cref{sec:loop_caustics} needs only Jacobian and Hessian products of the drift.
A loop-corrected deterministic sampler, the drift ODE supplemented by the $\mathcal{O}(g^2)$ correction, then delivers stochastic-process observables from two auxiliary linear equations solved along the same trajectory, at no stochastic-sampling cost. Whether that correction survives contact with a learned score, where the score error of \cref{sec:score_insertions} competes with the fluctuation term, is the first question to settle.
The insertion analysis of \cref{sec:score_insertions} adds two further tests. The response weighting derived there can compete against empirically tuned loss weightings, and the loop integrand, which identifies where along the trajectory the deterministic sampler deviates most, can allocate a limited budget of stochastic steps where they buy the most accuracy.

The power-counting scheme of \cref{sec:eft_power_counting} suggests a complementary program on the architecture side. 
Building generative models from the leading equivariant drift of \cref{sec:eft_permutation}, measuring the fitted interaction couplings, and testing the predicted hierarchy between operator degrees confront the expansion with practice. 
For data whose complexity exceeds the reach of a polynomial truncation, the EFT drift provides a symmetry-exact analytic backbone, with a small network learning only the higher-degree residual whose size the power counting bounds. 
Because the operator basis is exhaustive at each degree, it also contains drift terms that standard architectures do not realize, and the sign structure of the coefficients controls whether the flow contracts toward the data manifold.

The formulation connects deterministic and stochastic generative models within one action and identifies the analytically solvable limits. 
Sampler accuracy is now a calculation, carried out and validated in \cref{sec:loop_corrections}. 
Score quality and architecture design are so far only organized, but turning that organization into numbers is the work this framework makes possible.

\subsection*{Code availability}

The code reproducing the numerical validations of \cref{sec:ou_loop_example} -- the Ornstein--Uhlenbeck free theory, the interacting cubic drift, and the $24$-dimensional equivariant drift, including the Fokker--Planck reference solver and the Euler--Maruyama simulations, is publicly available at \url{https://github.com/ramonpeter/generative-path-integrals}.

\subsection*{Acknowledgements}

I thank Rafael Aoude for detailed feedback on the manuscript and for pointing me to the worldline effective field theory literature, and Claudius Krause for his questions about adversarial models that turned into a subsection of its own. I am grateful to Sofia Palacios Schweitzer for a detailed round of comments that improved the manuscript throughout, and to Nina Elmer, who was involved in the early days of this project, and Henning Bahl for reading it and for their comments along the way.

\clearpage
\bibliographystyle{tepml}
\bibliography{refs}
\end{document}

%% file: incl_settings.tex
\usepackage[utf8]{inputenc} 
\usepackage[T1]{fontenc} 	
\usepackage[english]{babel} 

\usepackage[bitstream-charter]{mathdesign}
\usepackage{pifont}
\usepackage[top=12mm,bottom=12mm,left=30mm,right=30mm,head=12mm,includeheadfoot,a4paper]{geometry}
\usepackage{mathtools} 		
\usepackage{float} 			
\usepackage{graphicx} 		
\usepackage{tabularx} 		
\usepackage{booktabs} 		
\usepackage{color, xcolor} 	
\definecolor{Rcolor}{HTML}{E99595}
\definecolor{Gcolor}{HTML}{C5E0B4}
\definecolor{Bcolor}{HTML}{9DC3E6}
\definecolor{Ycolor}{HTML}{FFE699}
\definecolor{Vcolor}{HTML}{CFB4E0}
\usepackage{pdfpages} 		
\usepackage{extarrows} 		
\usepackage{multirow} 		
\usepackage{multicol} 		
\usepackage{enumitem} 		
\usepackage{xspace} 		
\usepackage{stackrel} 		
\usepackage{tikz} 			
\usetikzlibrary{arrows.meta, positioning, fit, calc}
\usepackage{braket} 		
\usepackage{bm} 			
\usepackage{tensor} 		
\usepackage{slashed} 		
\usepackage{siunitx} 		
\usepackage{lastpage} 		
\usepackage{cite} 			
\usepackage[normalem]{ulem} 
\usepackage{fontawesome} 	
\usepackage{tocloft} 		
\usepackage{titlesec} 		
\usepackage{doi} 			
\usepackage{hyperref} 		
\usepackage[most]{tcolorbox} 					
\usepackage[nameinlink, capitalize]{cleveref} 	
\usepackage[nottoc, notlot, notlof]{tocbibind} 	
\usepackage[ruled, vlined]{algorithm2e} 		
\usepackage{makecell}
\usepackage[makeroom]{cancel}
\usepackage{accents}
\usepackage{adjustbox}

\numberwithin{equation}{section}

\makeatletter
\def\BState{\State\hskip-\ALG@thistlm}
\makeatother

\makeatletter
\@ifundefined{pdfoutput}{}{\DeclareGraphicsRule{*}{mps}{*}{}}
\makeatother

\makeatletter
\DeclareRobustCommand*{\bfseries}{%
   \not@math@alphabet\bfseries\mathbf
   \fontseries\bfdefault\selectfont
   \boldmath
}
\makeatother

\DeclareSymbolFont{usualmathcal}{OMS}{cmsy}{m}{n}
\DeclareSymbolFontAlphabet{\mathcal}{usualmathcal}

\SetArgSty{textnormal}
\SetKwComment{Comment}{{\small\#}~}{}
\SetCommentSty{mycommfont}

\setitemize{itemsep=2pt,topsep=2pt,parsep=0pt,partopsep=0pt,leftmargin=*}
\setenumerate{itemsep=0pt,topsep=2pt,parsep=0pt,partopsep=0pt,labelindent=3pt,leftmargin=*}
\usepackage{diagbox} 

%% file: incl_shortcuts.tex
\definecolor{red_cb}{HTML}{e41a1c}
\definecolor{blue_cb}{HTML}{377eb8}
\definecolor{green_cb}{HTML}{4daf4a}
\definecolor{purple_cb}{HTML}{984ea3}
\definecolor{orange_cb}{HTML}{ff7f00}

\definecolor{EmeraldGreen}{HTML}{1ea78d}
\definecolor{EnglishRed}{HTML}{b02427}
\definecolor{EmeraldGreen}{HTML}{1ea78d}
\definecolor{EnglishRed}{HTML}{b02427}
\hypersetup{
	pdftitle={Unifying Generative Models with Path Integrals},
	pdfauthor={Ramon Winterhalder},
	colorlinks=true, 			
	linkcolor={red!50!black}, 	
	citecolor=EmeraldGreen, 	
	urlcolor=EmeraldGreen 	
}

\newcommand{\eg}{\text{e.g.}\;}
\newcommand{\ie}{\text{i.e.}\;}

\newcommand{\eqcomma}{\;,} 	
\newcommand{\eqperiod}{\;.} 	

\newcommand{\mwith}{\text{with}}
\newcommand{\mand}{\text{and}}
\newcommand{\mfor}{\text{for}}

\newcommand{\group}{g}

\def\d{\mathrm{d}}

\newcommand\one{\leavevmode\hbox{\small1\normalsize\kern-.33em1}}
\newcommand{\tr}{\operatorname{Tr}}			
\newcommand{\imag}{\mathrm{i}} 				

\newcommand{\arXiv}[2][]{%
	\ifthenelse{\equal{#1}{}}%
	{\href{http://arxiv.org/abs/#2}{arXiv:#2}}%
	{\href{http://arxiv.org/abs/#2}{arXiv:#2~[#1]}}}

\def\slashchar#1{\setbox0=\hbox{$#1$}           
   \dimen0=\wd0                                 
   \setbox1=\hbox{/} \dimen1=\wd1               
   \ifdim\dimen0>\dimen1                        
      \rlap{\hbox to \dimen0{\hfil/\hfil}}      
      #1                                        
   \else                                        
      \rlap{\hbox to \dimen1{\hfil$#1$\hfil}}   
      /                                         
   \fi}

\newcommand{\tikznode}[2]{%
\ifmmode%
\tikz[remember picture,baseline=(#1.base),inner sep=0pt] \node (#1) {$#2$};%
\else
\tikz[remember picture,baseline=(#1.base),inner sep=0pt] \node (#1) {#2};%
\fi}

\def\mathswitchr#1{\relax\ifmmode{\mathrm{#1}}\else$\mathrm{#1}$\xspace\fi}
\def\mathswitch#1{\relax\ifmmode#1\else$#1$\xspace\fi}

%% file: tikz_setup.tex
\usetikzlibrary{calc}
\usetikzlibrary{decorations}
\usetikzlibrary{decorations.pathreplacing}
\usetikzlibrary{decorations.pathmorphing}
\usetikzlibrary{decorations.markings}
\usetikzlibrary{external}
\usetikzlibrary{fit}
\usetikzlibrary{graphs}
\usetikzlibrary{patterns}
\usetikzlibrary{positioning}
\usetikzlibrary{shapes}

\tikzset{/pgf/decoration/.cd,angle step/.initial=3}

\newdimen\tmpdimen
\pgfdeclaredecoration{complete sines}{initial}{
	\state{initial}[
		width=+0pt,
		next state=move,
		persistent precomputation={%
			\pgfmathparse{frac(\pgfdecoratedremainingdistance / \pgfdecorationsegmentlength)}%
			\let\bastardlength=\pgfmathresult%
			\pgfmathparse{\pgfdecoratedremainingdistance / (\pgfdecoratedremainingdistance / \pgfdecorationsegmentlength - \bastardlength)}%
			\let\pgfdecorationsegmentlength=\pgfmathresult%
			\pgfmathparse{\pgfkeysvalueof{/pgf/decoration/angle step}}%
			\let\anglestep=\pgfmathresult%
			\let\currentangle=\pgfmathresult%
			\pgfmathsetlengthmacro{\pointsperanglestep}{\pgfdecorationsegmentlength / 360.0 * \anglestep}%
		}] {}
	\state{move}[width=+\pointsperanglestep, next state=draw]{
		\pgfpathmoveto{\pgfpointorigin}
	}
	\state{draw}[
		width=+\pointsperanglestep,
		switch if less than=1.25*\pointsperanglestep to final, 
		persistent postcomputation={%
			\pgfmathparse{mod(\currentangle+\anglestep, 360)}%
			\let\currentangle=\pgfmathresult%
	}]{%
		\pgfmathsin{+\currentangle}%
		\tmpdimen=\pgfdecorationsegmentamplitude%
		\tmpdimen=\pgfmathresult\tmpdimen%
		\divide\tmpdimen by2\relax%
		\pgfpathlineto{\pgfqpoint{0pt}{\tmpdimen}}%
	}
	\state{final}{
		\ifdim\pgfdecoratedremainingdistance>0pt\relax
			\pgfpathlineto{\pgfpointdecoratedpathlast}
		\fi
	}
}

\makeatletter
\def\pgf@lib@dec@parsenum#1{%
    \gdef\pgf@lib@dec@computed@width{0 pt}%
    \tsx@pgf@lib@dec@parsenum#1+endmarker+%
    \ifdim\pgf@lib@dec@computed@width<0pt\relax%
        \pgfmathparse{\pgfdecoratedpathlength\pgf@lib@dec@computed@width}
        \edef\pgf@lib@dec@computed@width{\pgfmathresult pt}%
    \fi%
}

\def\tsx@pgf@lib@dec@parsenum@endmarker{endmarker}

\def\tsx@pgf@lib@dec@parsenum#1+{
    \def\temp{#1}%
    \ifx\temp\tsx@pgf@lib@dec@parsenum@endmarker%
    \else%
        \tsx@pgf@lib@dec@parsenum@one{#1}%
        \expandafter\tsx@pgf@lib@dec@parsenum%
    \fi%
}

\def\tsx@pgf@lib@dec@parsenum@one#1{%
  \pgfmathparse{#1}%
  \ifpgfmathunitsdeclared%
    \pgfmathparse{\pgf@lib@dec@computed@width + \pgfmathresult pt}%
  \else%
    \pgfmathparse{\pgf@lib@dec@computed@width + \pgfmathresult*\pgfdecoratedpathlength*1pt}%
  \fi%
  \edef\pgf@lib@dec@computed@width{\pgfmathresult pt}%
}
\makeatother

\tikzset{
    vertex/.style={
        circle,
        fill,
        inner sep=0,
        minimum size=3,
        outer sep=-1
    },
    fermion/.style={
        decoration={
            markings,
            mark=at position 0.5 +3.4pt with \arrow{>}
        },
        postaction=decorate,
        thick
    },
    blob/.style={
        circle,
        draw,
        outer sep=-0.2,
        pattern color=gray,
        pattern=crosshatch,
        thick
    },
	coupling/.style={
		circle,
		draw,
		inner sep=2.,
		outer sep=-0.1,
		fill=gray!50!white,
		thick
	},
	blobcorr/.style={
		circle,
		draw,
		outer sep=-0.2,
		fill=gray,
		thick
	},
    vector/.style={
        decorate=true,
        decoration={
            complete sines,
            segment length=6pt,
            amplitude=6pt
        },
        thick
    },
    gluon/.style={
        decorate=true,
        decoration={
            complete sines,
            segment length=0.2em,
            amplitude=0.25em
        }
    }
}